\documentclass[10pt,twocolumn,letterpaper]{article}

\usepackage[pagenumbers]{cvprnotimes}

\usepackage[dvipsnames]{xcolor}
\usepackage{graphicx}
\usepackage{amsmath}
\usepackage{amssymb}
\usepackage{booktabs}
\usepackage{subcaption}
\usepackage{bbm}

\usepackage[pagebackref,breaklinks,colorlinks]{hyperref}

\usepackage[capitalize]{cleveref}
\crefname{section}{Sec.}{Secs.}
\Crefname{section}{Section}{Sections}
\Crefname{table}{Table}{Tables}
\crefname{table}{Tab.}{Tabs.}

\def\E{\mathbb{E}}
\newcommand{\mc}[1]{\mathcal{#1}}
\renewcommand{\vec}[1]{{\boldsymbol{#1}}}
\DeclareMathOperator*{\argmax}{arg\,max}

\def\cls{C}
\newcommand{\cl}[1]{C_{#1}}
\def\totalcl{K}
\newcommand{\pool}[1]{\mc{P}_{#1}}
\def\poolsize{N_{\pool{}}}
\def\poolsteps{S_{\pool{}}}
\newcommand{\dataset}[1]{\mc{D}_{#1}}
\newcommand{\model}[1]{\mc{M}_{#1}}
\newcommand{\task}[1]{\mc{T}_{#1}}
\def\pubdataset{\mc{D}_*}
\newcommand{\loss}[1]{\mc{L}_{#1}}
\def\confidence{\Lambda}
\def\samples{S}
\newcommand{\fraction}[1]{\gamma_{#1}}
\newcommand{\labels}[1]{\ell_{#1}}
\def\onelabel{l}
\def\numcheckpoints{\Delta}

\def\accpriv{\beta_{\rm priv}}
\def\accsh{\beta_{\rm sh}}

\def\regemb{\nu_{\rm emb}}
\def\regaux{\nu_{\rm aux}}

\def\step{t}
\newcommand{\graph}[1]{\mc{G}_{#1}}
\newcommand{\edges}[1]{\mc{E}_{#1}}

\def\imagenet{ImageNet}
\def\skew{s}

\begin{document}

\title{Decentralized Learning with Multi-Headed Distillation}

\author{Andrey Zhmoginov \quad Mark Sandler \quad Nolan Miller \quad Gus Kristiansen \quad Max Vladymyrov \vspace{.1cm} \\
Google Research\\
% 1600 Amphitheatre Pkwy \\ Mountain View, CA 94043, USA \\
{\tt\small \{azhmogin,sandler,namiller,gusatb,mxv\}@google.com}
}

\maketitle

% =================================================================================
%   ABSTRACT
% =================================================================================

\begin{abstract}
Decentralized learning with private data is a central problem in machine learning. We propose a novel distillation-based decentralized learning technique that allows multiple agents with private non-iid data to learn from each other, without having to share their data, weights or weight updates. Our approach is communication efficient, utilizes an unlabeled public dataset and uses multiple auxiliary heads for each client, greatly improving training efficiency in the case of heterogeneous data. This approach allows individual models to preserve and enhance performance on their private tasks while also dramatically improving their performance on the global aggregated data distribution. We study the effects of data and model architecture heterogeneity and the impact of the underlying communication graph topology on learning efficiency and show that our agents can significantly improve their performance compared to learning in isolation.
\end{abstract}

% =================================================================================
%   INTRODUCTION
% =================================================================================

\section{Introduction}

    Supervised training of large models historically relied on access to massive amounts of labeled data.
    Unfortunately, since data collection and labeling are very time-consuming, curating new high-quality datasets remains expensive and practitioners are frequently forced to get by with a limited set of available labeled datasets.
    Recently it has been proposed to circumvent this issue by utilizing the existence of large amounts of siloed private information.
    Algorithms capable of training models on the entire available data without having a direct access to private information have been developed with Federated Learning approaches \cite{McMahan17} taking the leading role.
    
    \begin{figure}[t]
        \centering
        \includegraphics[width=0.35\textwidth]{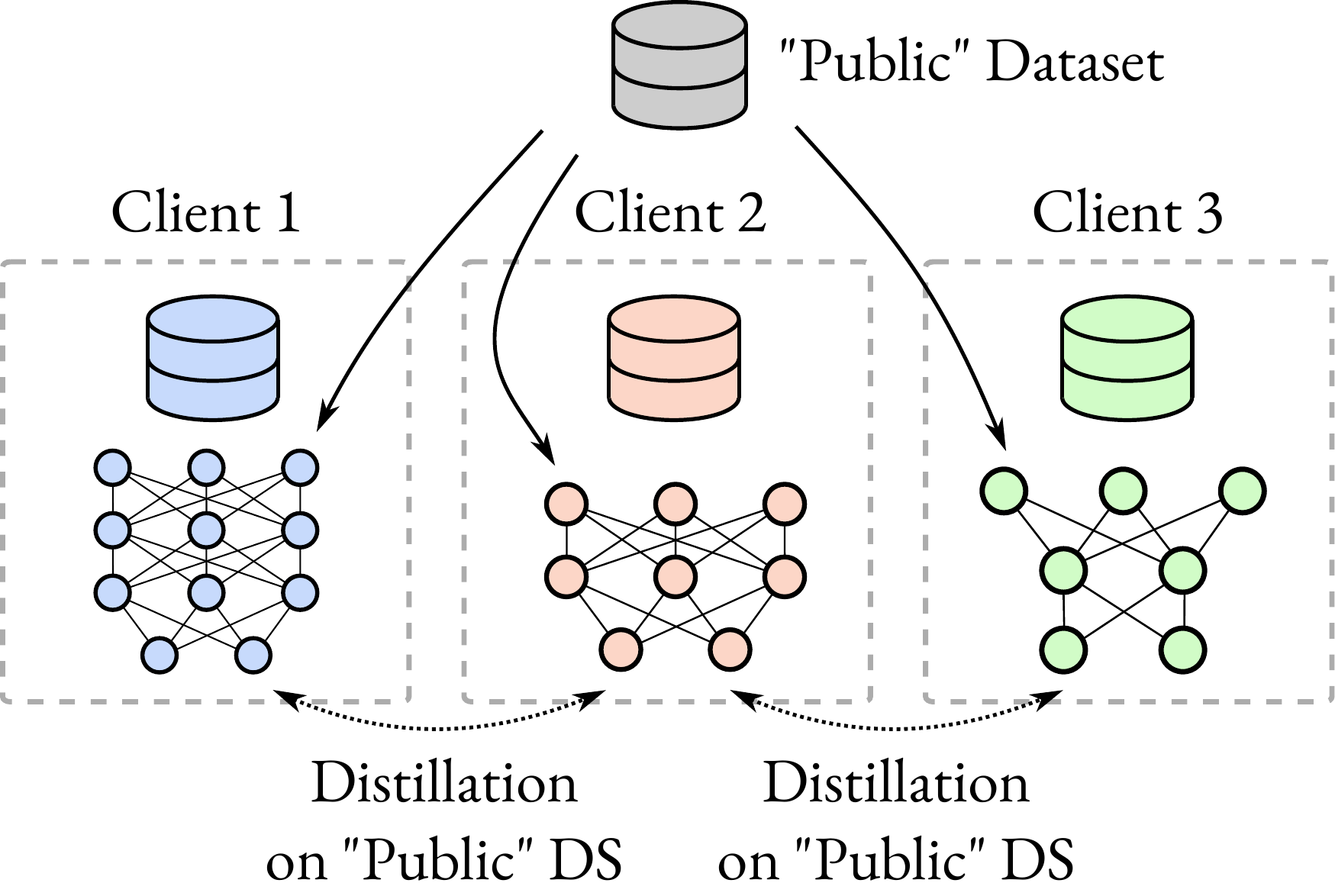}
        \caption{
            Conceptual diagram of a distillation in a distributed system.
            Clients use a public dataset to distill knowledge from other clients, each having their primary private dataset.
            Individual clients may have different architectures and different objective functions.
        }
        \label{fig:main}
    \end{figure}
    
    While very effective in large-scale distributed environments, more canonical techniques based on federated averaging, have several noticeable drawbacks.
    First, gradient aggregation requires individual models to have fully compatible weight spaces and thus identical architectures.
    While this condition may not be difficult to satisfy for sufficiently small models trained across devices with compatible hardware limitations, this restriction may be disadvantageous in a more general setting, where some participant hardware can be significantly more powerful than the others.
    Secondly, federated averaging methods are generally trained in a centralized fashion.
    Among other things, this prohibits the use of complex distributed communication patterns and implies that different groups of clients cannot generally be trained in isolation from each other for prolonged periods of time.
    
    Another branch of learning methods suitable for distributed model training on private data are those based on distillation \cite{Bucila06,Ba14,Hinton15}.
    Instead of synchronizing the inner states of the models, such methods use outputs or intermediate representations of the models to exchange the information.
    The source of data for computing exchanged model predictions is generally assumed to be provided in the form of publicly available datasets \cite{Guha19} that do not have to be annotated since the source of annotation can come from other models in the ensemble (see Figure~\ref{fig:main}).
    One interesting interpretation of model distillation is to view it as a way of using queries from the public dataset to indirectly gather information about the weights of the network (see Appendix~\ref{app:theory}).
    Unlike canonical federated-based techniques, where the entire model state update is communicated, distillation only reveals {\em activations} on specific samples, thus potentially reducing the amount of communicated bits of information.
    By the data processing inequality, such reduction, also translates into additional insulation of the private data used to train the model from adversaries.
    However, it is worth noting that there exists multiple secure aggregation protocols including SecAgg \cite{Bonawitz17} that provide data privacy guarantees for different Federated Learning techniques.
    
    The family of approaches based on distillation is less restrictive than canonical federated-based approaches with respect to the communication pattern, supporting fully distributed knowledge exchange.
    It also permits different models to have entirely different architectures as long as their outputs or representations are compatible with each other.
    It even allows different models to use various data modalities and be optimizing different objectives, for example mixing supervised and self-supervised tasks within the same domain.
    Finally, notice that the distillation approaches can and frequently are used in conjunction with weight aggregation \cite{Lin20,Shen20,Sturluson21,Wu21}, where some of the participating clients may in fact be entire ensemble of models with identical architectures continuously synchronized using federated aggregation (see Figure~\ref{fig:main_updated} in Supplementary).
    
    \paragraph{Our contributions.} In this paper, we propose and empirically study a novel distillation-based technique that we call Multi-Headed Distillation (MHD) for distributed learning on a large-scale \imagenet{} \cite{Deng2009} dataset.
    Our approach is based on two ideas: (a) inspired by self-distillation \cite{Furlanello18,Yang19,Ahn19} we utilize multiple model heads distilling to each other (see Figure~\ref{fig:aux}) and (b) during training we simultaneously distill client model predictions and intermediate network embeddings to those of a target model.
    These techniques allow individual clients to effectively absorb more knowledge from other participants, achieving a much higher accuracy on a set of all available client tasks compared with the naive distillation method.
    
    In our experiments, we explore several key properties of the proposed model including those that are specific to decentralized distillation-based techniques.
    First, we analyse the effects of data heterogeneity, studying two scenarios in which individual client tasks are either identical or very dissimilar.
    We then investigate the effects of working with nontrivial communication graphs and using heterogeneous model architectures.
    Studying complex communication patterns, we discover that even if two clients in the ensemble cannot communicate directly, they can still learn from each other via a chain of interconnected clients.
    This ``transitive'' property relies in large part on utilization of multiple auxiliary heads in our method.
    We also conduct experiments with multi-client systems consisting of both ResNet-18 and ResNet-34 models \cite{He16} and demonstrate that: (a) smaller models benefit from having large models in the ensemble, (b) large models learning from a collection of small models can reach higher accuracies than those achievable with small models only.

% =================================================================================
%   RELATED WORK
% =================================================================================

\section{Related Work}

    \paragraph{Personalized Federated Learning.}
    While many early canonical Federated Learning approaches trained a single global model for all clients \cite{McMahan17}, it has been quickly realized that non-IID nature of private data in real systems may pose a problem and requires personalized approaches \cite{Li20}.
    Since then many Personalized Federated Learning approaches have been developed, many covered in the surveys \cite{Kulkarni20,Tan21}.

    \paragraph{Federated Distillation.}
    Emergence of Federated Distillation was motivated by the need to perform learning across ensembles of heterogeneous models\footnote{note that multiple existing approaches like \cite{Pillutla22,Wan22} allow using \mbox{FedAvg} for training heterogeneous model ensembles}, reducing communication costs and improving performance on non-IID data.
    Existing distillation-based approaches can be categorized based on the system setup and the types of the messages passed between participants.
    A number of approaches including \cite{Makhija22,Guha19,Lin20,Shen20,Chen21,Sturluson21,Zhu21,Wu21} combine aggregation of weight updates with model distillation.
    They are typically centralized and frequently involve client-side distillation, which may restrict the size of the aggregated model.
    A different body of work is concentrated on centralized systems, where only model predictions are communicated between the clients and the server \cite{Li19,Itahara20,He20,Sattler20,Sun21,Zhang21,Gong22,Nguyen22}.
    Another related family of approaches is based on communicating embedding prototypes \cite{Tan22}, or using embeddings for distillation directly \cite{Aguilar20,Nguyen22}.
    In this paper, we concentrate on a more general decentralized setup, where there is not single central authority and all clients exchange knowledge via distillation \cite{Bistritz20}.

% =================================================================================
%   MODEL
% =================================================================================

\section{Model}

    \subsection{Setup}
    \label{sec:setup}
    
        We consider a system of $\totalcl$ {\em clients} $\cls=\{\cl{1},\dots,\cl{\totalcl}\}$.
        Each client $\cl{i}$ is assumed to possess their own {\em private dataset} $\dataset{i}$ while training a {\em private model} $\model{i}$ that solves a corresponding task $\task{i}$.
        In the following, we assume that all tasks $\task{i}$ are supervised.
        
        While using their local dataset $\dataset{i}$ to train the private model, each client can also communicate with other clients to learn from them.
        At each global training step $\step$, we define a local directed graph $\graph{\step}$ that determines the pattern of this communication.
        While the set of nodes of $\graph{\step}$ is fixed to be the set of all clients, the set of edges $\edges{\step}$ with the corresponding incidence function can be dynamic and change every training step.
        
        The local datasets $\dataset{i}$ are not directly exchanged between the clients, instead the information exchange occurs via a shared {\em public} source of unlabeled data $\pubdataset$.
        We assume that at training step $\step$, each client $\cl{i}$ can perform inference on a set of public samples and request the results of a similar computation on the same samples from other clients that are incident to it by directed edges of $\graph{\step}$.
        In other words, each client $\cl{i}$ is optimizing a local objective $\loss{i}$ defined as:
        \begin{gather}
            \loss{i}(\step) = \loss{i,{\rm CE}} +
            \sum_\alpha \E_{x\sim \pubdataset} \loss{\rm dist}^{\alpha}(\psi^\alpha_i(x), \Phi_{t,i}^\alpha),
        \end{gather}
        where $\loss{i,{\rm CE}} \equiv \E_{(x,y)\sim \dataset{i}} \loss{\rm CE}(x,y)$ and $\loss{\rm CE}$ is a cross-entropy loss optimized locally by each client on their private data $\dataset{i}$, $\loss{\rm dist}^{\alpha}$ is a collection of different {\em distillation losses} enumerated by $\alpha$ that use some {\em local} computation result $\psi^\alpha_i$ and a {\em remote} results $\Phi_{t,i}^\alpha(x) \equiv \{ \phi^\alpha_j(x) | j\in e_\step(i) \}$ computed on the same sample and $e_\step(i)$ is a set of clients connected to $i$ via a set of outgoing edges (from $\graph{\step}$).
        
        Notice that in contrast with Federated Learning, here we do not require different models $\model{i}$ to have compatible architectures, but instead optimize local and remote sample representations $\psi_i(x)$ and $\phi_j(x)$ to be compatible.
        In the next section, we discuss several potential choices of the distillation losses.
        
        In this paper, we are interested in evaluating the impact that the communication and cross-learning between the clients has on (a) how well these models can be suited for their original private tasks and (b) how much of the knowledge gets shared and distributed to the other tasks over time.
        Notice that if each client has a sufficiently simple model and enough training data (making the model underfit), the communication between individual models is not expected to improve their private task performance, but can only enhance their learned representations making them more suitable for adapting to other client's tasks.
        However, if the private training data is scarce (making the model overfit), the model communication could improve generalization and ultimately improve client performance on their private tasks.
    
    \subsection{Distillation Losses}
    \label{sec:losses}
    
        \paragraph{Embedding distillation.}
        We utilize the embedding regularization loss \cite{Aguilar20,Nguyen22} in our experiments.
        If $\xi_i(x)$ is an intermediate embedding produced for a sample $x$ coming from the shared public dataset by the model $\model{i}$, then we can choose $\psi^{\rm emb}_i(x) \equiv \xi_i(x)$, $\phi^{\rm emb}_j(x) \equiv \xi_j(x)$ and define $\loss{\rm dist}^{\rm emb}\left(\psi_i^{\rm emb}(x),\Phi_{t,i}^{\rm emb}(x)\right)$ as
        \begin{gather}
        \regemb \sum_{j\in e_t(i)} \rho\left(\|\psi_i^{\rm emb}(x)-\phi_j^{\rm emb}(x)\|\right),
        \end{gather}
        or simply $\regemb \sum_{j\in e_t(i)} \rho\left( \|\xi_i(x)-\xi_j(x)\|\right)$, where $\regemb$ is the weighting constant and $\rho(x)\in C^\infty$ is some monotonically growing function.
        The choice of this distillation loss forces compatibility between sample embeddings across the ensemble.
        In practice, we noticed that the embedding norms of different models frequently diverge during training, and to adapt to that we use normalized embeddings preserving regularization consistency across the entire duration of training: $\psi^{\rm norm}_i(x) \equiv \xi_i(x) / \|\xi_i(x)\|$.
        
        \paragraph{Prediction distillation.}
        Ability to predict on classes that are rarely present in private data can be improved by utilizing prediction vector as an additional distillation target.
        However, since $\model{i}$ is tasked with fitting ground truth on a particular dataset $\dataset{i}$, distilling this prediction to labels relevant for another client may be damaging for the model performance on $\task{i}$.
        Instead, we choose to add another single prediction head to $\model{i}$ that is distilled from all existing tasks thus (a) not polluting the main prediction head of the model $\model{i}$, but (b) at the same time forcing the intermediate representation $\xi_i(x)$ to contain information relevant for solving all existing tasks $\{\task{j}|j\in 1,\dots,\totalcl\}$.
        
        Let $\vec{h}_i(\xi_i(x))$ be the main head of the model $\model{i}$ used for computing $\loss{\rm CE}$ and $\vec{h}^{\rm aux}_i(\xi_i(x))$ be the auxiliary head.
        Then, the na\"ive prediction distillation loss takes the following form:
        \begin{gather}
            \label{eq:dist-aux-naive}
            \loss{\rm dist}^{\rm aux}[\vec{h}^{\rm aux},\vec{h}] \equiv - \regaux \sum_{j\in e_t(i)} \vec{h}_j \log \vec{h}^{\rm aux}_i(x),
        \end{gather}
        where $\regaux$ is the auxiliary loss weight.
        Here all the distillation targets from $e_t(i)$ are essentially treated the same irrespective of their confidence in their prediction.
        One way of integrating the knowledge of the distillation target quality is to use some {\em confidence} metric for their prediction on $x$.
        For example, we could consider the following modification of the loss \eqref{eq:dist-aux-naive}:
        \begin{gather}
            - \regaux \sum_{j\in e_t(i) \cup \{i\}} Q\left[ \confidence(\vec{h}_j); H[\vec{h}]\right] \times \vec{h}_j \log \vec{h}^{\rm aux}_i(x),
            \label{eq:dist-aux}
        \end{gather}
        where $\confidence(\vec{h}(x))$ is the confidence of the classifier prediction, $Q$ is some function of the client confidence and $H[\vec{h}] \equiv \{ \confidence(\vec{h}_j) | j\in e_t(i) \cup \{i\}\}$ is the information about confidence of all possible distillation targets including the $i^{\rm th}$ client itself.
        We considered perhaps the simplest choice for $\confidence$ defining it as $\argmax_k h_k(x)$.
        This measure of the model confidence that we end up using in our method is, of course, not reliable (see Appendix~\ref{app:theory}) and using a separate per-client density model $\rho_i(x)$ for detecting in-distribution and out-of-distribution samples could potentially improve model performance (for an alternative approach see \cite{Ma20}).
        For $Q$, we only considered perhaps the most obvious choice of $Q[\confidence(\vec{h}_j)]=1$ if $j^{\rm th}$ client has the largest confidence from $H$ and $0$ otherwise, effectively selecting the most confident client and using it as the distillation target (see Appendix~\ref{app:theory} for a detailed discussion).
    
        \begin{figure}[t]
            \centering
            \includegraphics[width=0.38\textwidth]{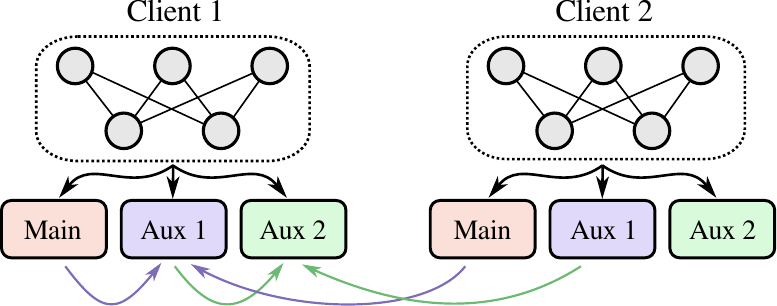}
            \caption{
                A pattern used for distilling multiple auxiliary heads.
                Here multiple auxiliary heads of ``Client 1'' are distilled from other auxiliary heads of the same model and from auxiliary heads of other clients (here ``Client 2'').
                Auxiliary head {\em Aux 1} is distilled from the main heads, auxiliary head {\em Aux 2} is distilled from auxiliary heads {\em Aux 1} and so on.
            }
            \label{fig:aux}
        \end{figure}
        
        \paragraph{Self-distillation with multiple auxiliary heads.}
        Self-distillation is a well-known technique that improves model performance by repeatedly using the previous iteration of the model as the distillation target for itself \cite{Furlanello18,Yang19,Ahn19,Mobahi20}.
        The most direct application of this technique to training an ensemble of models is to perform multiple cycles of self-distillation across all available networks.
        Here, however, we propose a different approach, where we modify a conventional training procedure by equipping each classifier with a collection of multiple auxiliary heads $\{\vec{h}^{{\rm aux},1}, \dots, \vec{h}^{{\rm aux},m}\}$.
        These auxiliary heads distill from each other by optimizing the following loss:
        \begin{equation}
            \label{eq:multiheads}
            \loss{\rm dist}^{\rm aux}[\vec{h}^{{\rm aux},1},\vec{h}] + 
            \sum_{k=2}^{m} \loss{\rm dist}^{\rm aux}[\vec{h}^{{\rm aux},k},\vec{h}^{{\rm aux},k-1}],
        \end{equation}
        where $\loss{\rm dist}^{\rm aux}[\vec{h}^{(a)},\vec{h}^{(b)}]$ is defined according to Eq.~\eqref{eq:dist-aux}.
        In other words, $\vec{h}^{{\rm aux},1}$ distills from $\vec{h}$ and $\vec{h}^{{\rm aux},k}$ distills from $\vec{h}^{{\rm aux},k-1}$ for all $1<k\le m$.
        This approach illustrated in Figure~\ref{fig:aux} is one of the core contributions of our paper.
    
        \paragraph{Communication efficiency.}
        In terms of communication efficiency, this approach could suffer from ineffective communication when the distillation targets are frequently a poor source of knowledge for a particular sample class.
        This problem would ideally require client awareness of the label distribution on each client that it communicates with.
        However, since in practice, prediction distillation (embedding distillation is more costly) only requires a transmission of several highest-confidence predictions for each sample, each step with batch size of $512$ would require a communication of only a few thousand floating point numbers (assuming that shared public set images could be uniquely identified with a small hash).
        At the same time, a single back-and-forth round of FedAvg communication of a ResNet-34 model would require more than $100$ million floating-point parameters, which would be equivalent to around $50{\rm k}$ prediction distillation steps.
    
    \subsection{Dataset}
    \label{sec:dataset}
    
        In this work, we study distributed learning in systems with varying degrees of data heterogeneity: from those where the distribution of data is the same across all clients, to more extreme cases where each client specializes on it's own unique task.
        We simulate these scenarios using an underlying labeled dataset $\dataset{}$.
        Let $\samples$ be the set of all samples from $\dataset{}$.
        Some fraction of samples $\fraction{\rm pub}$ (typically around $10\%$) is treated as a set of unlabeled {\em public} samples.
        The remaining samples are treated as the source of private data and are distributed without repetition across all of $\totalcl$ clients as discussed below.
        
        \paragraph{Label assignment.}
        Each client $\cl{i}$ is assigned a subset $\labels{i}$ of all labels, which are treated as {\em primary labels} for $\cl{i}$.
        Remaining labels from $\dataset{}$ not belonging to $\labels{i}$ are treated as {\em secondary labels} for $\cl{i}$.
        For each label $\onelabel$, we take all available samples and randomly distribute them across all clients.
        The probability of assigning a sample with label $\onelabel$ to a client $\cl{i}$ is chosen to be $1+\skew$ times higher for clients that have $\onelabel$ as their primary label.
        We call the parameter $\skew$ {\em dataset skewness}.
        As a result, in the iid case with $\skew=0$ all samples are equally likely to be assigned to any one of the clients.
        However, in the non-iid case in the limit of $\skew\to \infty$, all samples for label $\onelabel$ are only distributed across clients for which $\onelabel$ is primary.
        
        We considered two choices for selecting the primary label sets for the clients.
        One choice (we refer to as {\em even}) is to subdivide the set of all labels in such a way that each label has exactly $m$ corresponding primary clients.
        Another choice (we refer to as {\em random}) is to randomly assign each client $\cl{i}$ a random fixed-size subset of all labels.
        This choice creates a variation in the number of primary clients for different labels, making it a less idealized and more realistic setup even in the limit of $\skew\to \infty$.
        For example, for \imagenet{} with $1000$ classes, if it is subdivided between $8$ clients each receiving $250$ random labels: (a) around $100$ labels will be distributed evenly across all clients (no primary clients), (b) around $270$ labels will have a single primary client, (c) around $310$ labels will have two primary clients, (d) around $210$ labels will have three primary clients and (e) around $110$ remaining labels will have $4$ or more primary clients.

% =================================================================================
%   EXPERIMENTS
% =================================================================================

\section{Experiments}

    \subsection{Experimental Framework}
    
        In most of our experiments, we used \imagenet{} dataset with samples distributed across multiple clients as discussed in Section~\ref{sec:dataset}.
        The public dataset used for distillation was chosen by selecting $\fraction{\rm pub}=10\%$ of all available training samples and the remaining $90\%$ were distributed across clients as private labeled data.
        We used both {\em random} and {\em even} label distribution strategies and considered two cases of $\skew=0$ and $\skew=100$ corresponding to homogeneous and heterogeneous task distributions correspondingly.
        In most of our experiments, unless indicated otherwise, we used ResNet-34 models as individual clients, trained $8$ clients and each was assigned $250$ primary labels at {\em random}.
        The models were typically trained for 60\,000 or 120\,000 steps with SGD with momentum, batch size of $512$, cosine learning rate decay and the initial learning rate of $0.1$ and momentum $0.9$.
        
        Our experimental platform was based on distillation losses outlined in Section~\ref{sec:losses}.
        However, being restricted by computational efficiency needed to run numerous experiments, we made several implementation choices that deviated from the general formulation of Section~\ref{sec:losses}.
        Most importantly, individual clients do not directly exchange their predictions on the public dataset, but instead each client $\cl{i}$ keeps a rolling pool $\pool{i}$ of $\poolsize$ model checkpoints.
        In most of our experiments, $\poolsize$ was chosen to be equal to the total number of clients in the system.
        Every step, each client $\cl{i}$ picks a $\numcheckpoints$ random checkpoints from $\pool{i}$ and uses them for performing a distillation step on a new batch.
        Each pool $\pool{i}$ is updated every $\poolsteps$ steps, when a new checkpoint for one of the other clients is added into the pool (replacing another random checkpoint).
        In most of our experiments, we used a single distillation client on every step, i.e., $\numcheckpoints=1$ and $e_t(i)$ defined in Sec.~\ref{sec:setup} contains a single element every step $t$.
        However, a separate exploration of the parameter $\numcheckpoints$ was also performed.
        Also, since in most of our experiments we used $\poolsteps=200$, infrequent pool updates would typically introduce a time lag causing the model to distill knowledge from somewhat outdated checkpoints.
    
        \begin{figure}[t]
            \centering
            \begin{subfigure}{0.45\textwidth}
            \begin{center}
            \includegraphics[width=.4\textwidth]{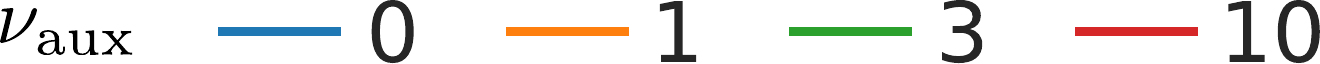}
            \vspace{3pt}
            \end{center}
            \end{subfigure}
            \begin{subfigure}{0.235\textwidth}
            \includegraphics[width=.95\textwidth]{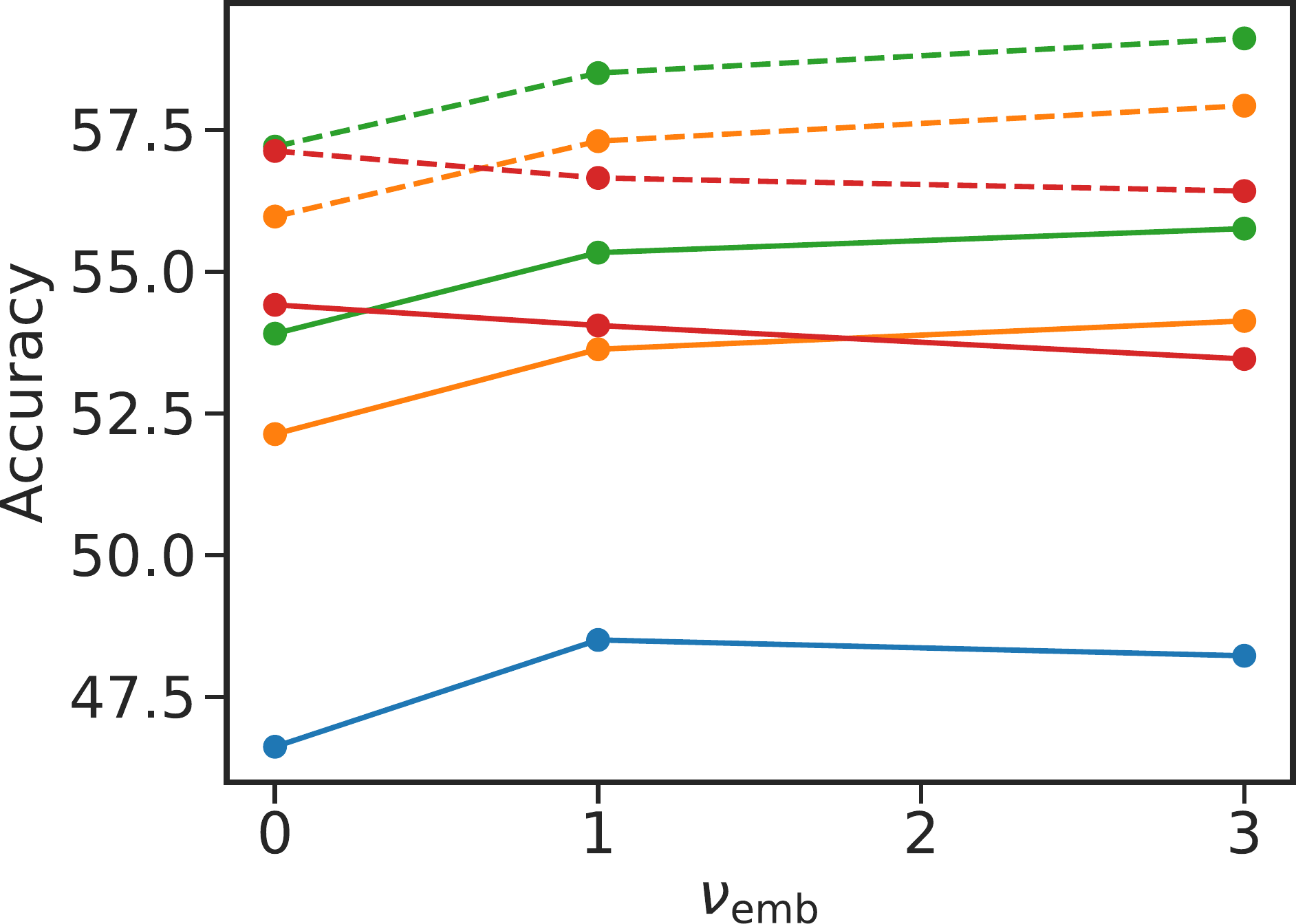}
            \caption{IID ($\skew=0$), Private Acc.}
            \end{subfigure}
            \begin{subfigure}{0.235\textwidth}
            \includegraphics[width=.95\textwidth]{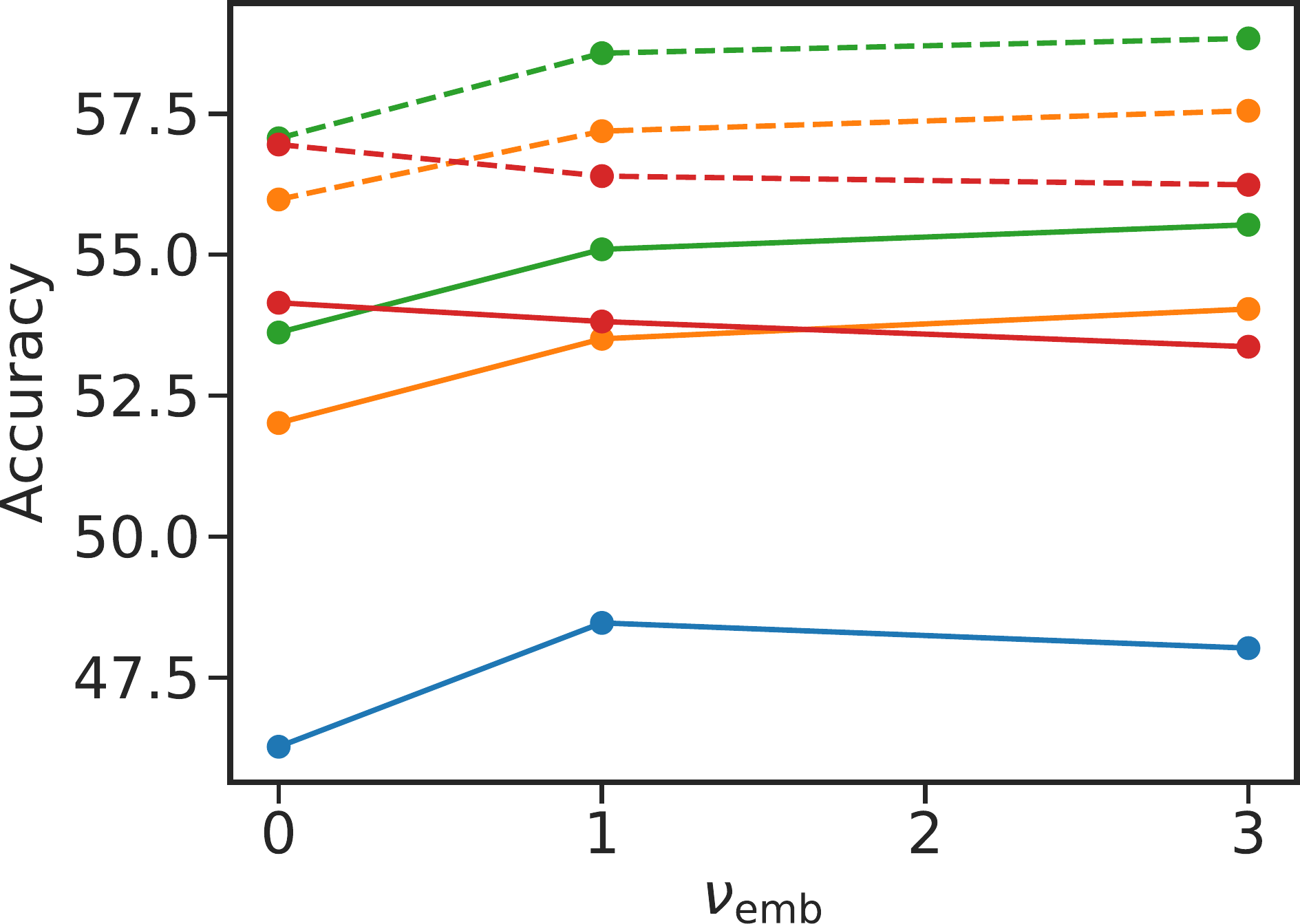}
            \caption{IID ($\skew=0$), Shared Acc.}
            \end{subfigure}
            \begin{subfigure}{0.235\textwidth}
            \includegraphics[width=.95\textwidth]{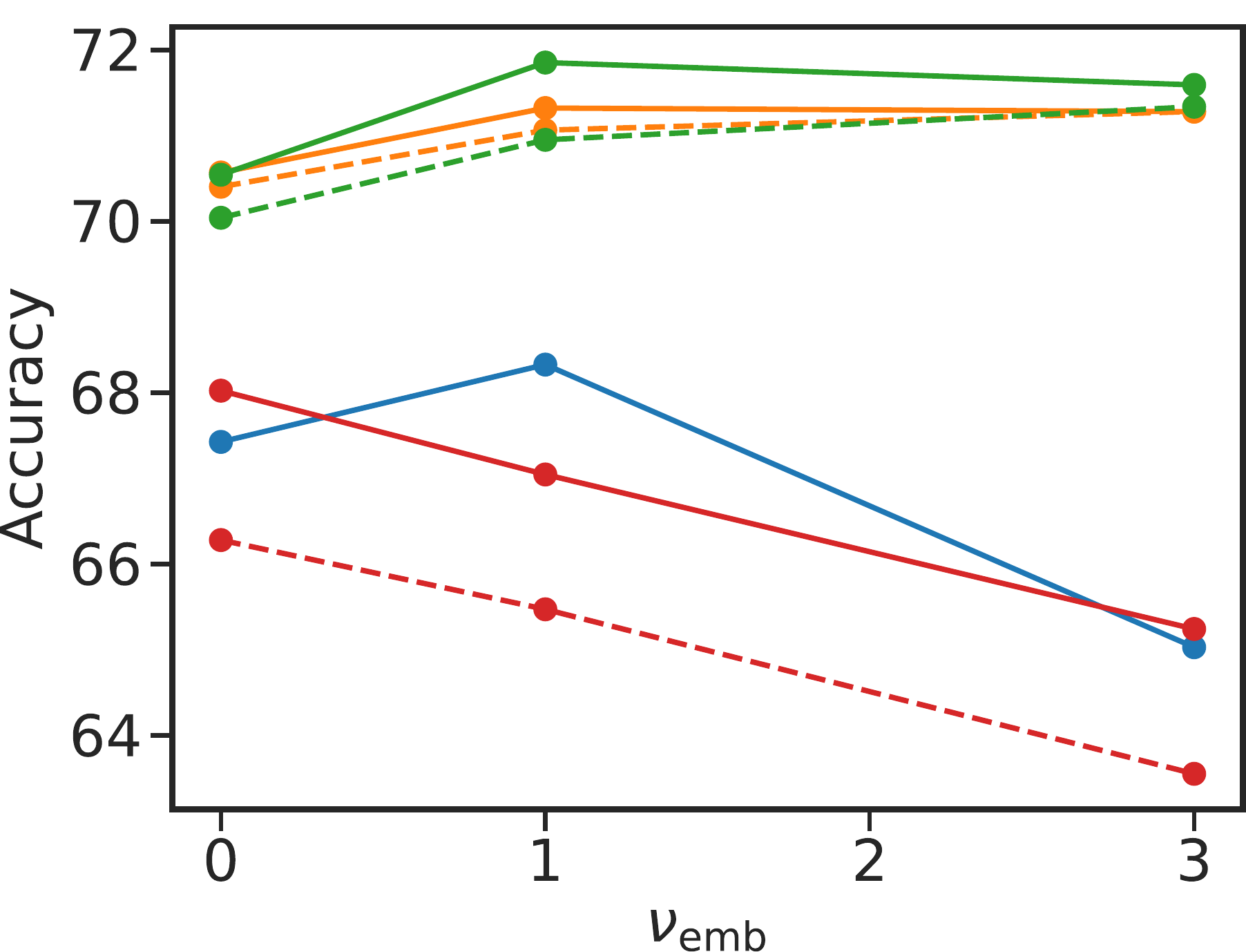}
            \caption{non-IID ($\skew=100$), Private Acc.}
            \end{subfigure}
            \begin{subfigure}{0.235\textwidth}
            \includegraphics[width=.95\textwidth]{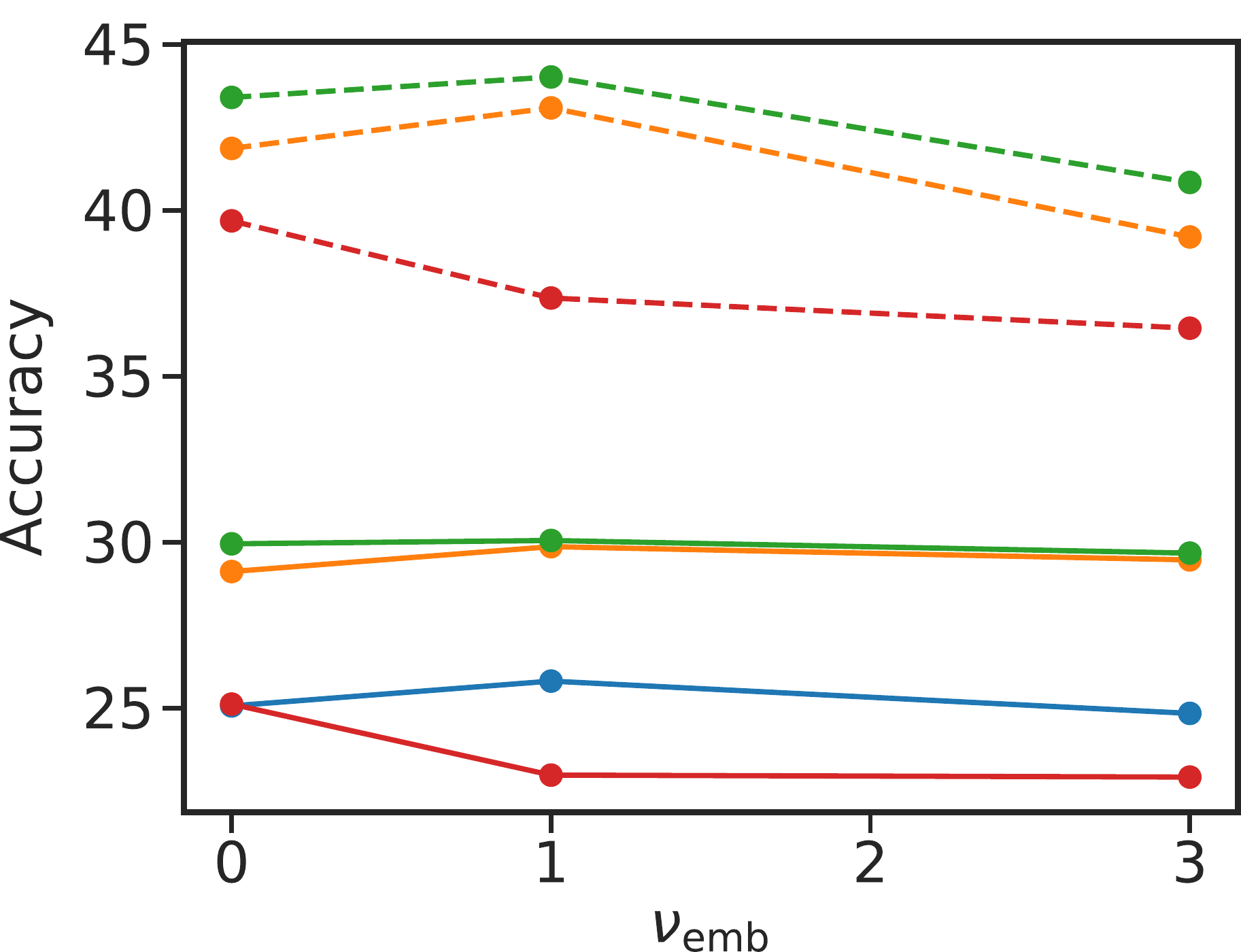}
            \caption{Non-IID ($\skew=100$), Shared Acc.}
            \end{subfigure}
            \caption{
                Comparison of {\em private} (on the client's dataset) and {\em shared} accuracies (on a uniform class distribution) for models trained on datasets with iid and non-iid distributions (see \cref{sec:dataset}) (a) with $\skew=0$ and (b) $\skew=100$.
                Both the main head (solid) and the auxiliary head accuracies (dashed) are shown.
                Four values of $\regaux$ are shown: $0.0$ ({\color{NavyBlue} blue}), $1.0$ ({\color{Orange} orange}), $3.0$ ({\color{Green} green}), $10.0$ ({\color{BrickRed} red}).
                The accuracies are seen to peak for $\regaux = 3$ and $\regemb = 3$ for $\skew=0$ and $\regemb = 1$ for $\skew=100$.
            }
            \label{fig:regularization-effect}
        \end{figure}
    
    \subsection{Embedding and Multi-Headed Distillation}
    \label{sec:exp-homo}
    
        In this section we start exploring distillation technique in the simplest scenario with identical model architectures and a complete graph connectivity, where each model can distill knowledge from any other existing client.
    
    \subsubsection{Evaluating Basic Distillation Approaches}
    \label{sec:exp-basic}
        Consider a set of models with identical ResNet-based architectures learning on their private subsets of \imagenet{} and distilling the knowledge from each other assuming a complete connectivity of the communication graph.
        Here we compare the efficiency of knowledge transfer for different distillation approaches: (a) distilling sample embeddings preceding the final logits layer ({\em embedding distillation}) and (b) distilling actual model predictions ({\em prediction distillation}) (see Sec.~\ref{sec:losses}).
        Specifically, we consider two extreme cases of an iid ($\skew=0$) and non-iid ($\skew=100$) distributed \imagenet{} datasets and study the final performance of individual agents while varying the strengths of the embedding and the prediction distillation losses, $\regemb$ and $\regaux$ correspondingly.
        
        In our experiments, we study the performance of primary and auxiliary model heads on two data distributions: (a) {\em private dataset} defining the primary problem that the client is tasked with and (b) {\em shared dataset} reflecting the uniform label distribution averaged across all clients.
        Any technique improving the private dataset accuracy $\accpriv$ can be viewed as successful at learning from other clients and translating the acquired knowledge into better performance on their own task.
        On the other hand, a technique improving the shared dataset accuracy $\accsh$ is successful at learning a more robust representation that can be easily adapted to solving other possible tasks (seen by other clients).
        Both of these potential capabilities can be viewed as positive outcomes of cross-client communication and learning, but their utility may be application specific.
        
        Figure~\ref{fig:regularization-effect} summarizes our empirical results (see Appendix B for raw numbers) showing the measurements of the average private accuracy $\accpriv$, that is the accuracy of each client on their respective dataset $\dataset{i}$, and the averaged shared accuracy $\accsh$ measured on a dataset with a uniform label distribution identical to that of the original \imagenet{}.
        While $\accpriv$ measures how well a particular client performs on their own task, $\accsh$ is a reflection of the world knowledge (some may be irrelevant for the private task) that the client learns from other participants.
    
        Figure~\ref{fig:regularization-effect} contains several interesting findings: (a) while both regularization techniques are useful for improving model performance, there is a threshold beyond which they start deteriorating both accuracies; (b) taken alone prediction distillation seems to have a stronger positive effect than the embedding distillation, while embedding distillation is more effective in the $\skew=0$ case; (c) however, the best results are obtained by combining both distillation techniques.
        Furthermore, we see that the distillation techniques generally improve both $\accpriv$ and $\accsh$ simultaneously.
        Notice that the positive effect of $\regaux$ suggests that training a separate auxiliary head has an effect on the model embedding that leads to an improved performance on the main head trained with the client's private dataset alone.
        Another interesting observation is that for uniform datasets with a small $\skew$, the auxiliary head ends up having better performance on both the private and shared tasks (identical for $\skew=0$).
        At the same time, in a non-iid dataset with $\skew=100$, auxiliary head performs much better on the shared dataset, but lags behind on the private task since it is not trained on it directly.
    
    \subsubsection{Improving Distillation Efficiency}
    \label{sec:improving}
    
        While Figure~\ref{fig:regularization-effect} shows a clear evidence that distillation techniques can be useful for distributed learning even in the case of heterogeneous client data, there is a room for further improvement.
    
        \begin{figure}
            \centering
            \begin{subfigure}[t]{0.23\textwidth}
            \includegraphics[width=0.97\textwidth]{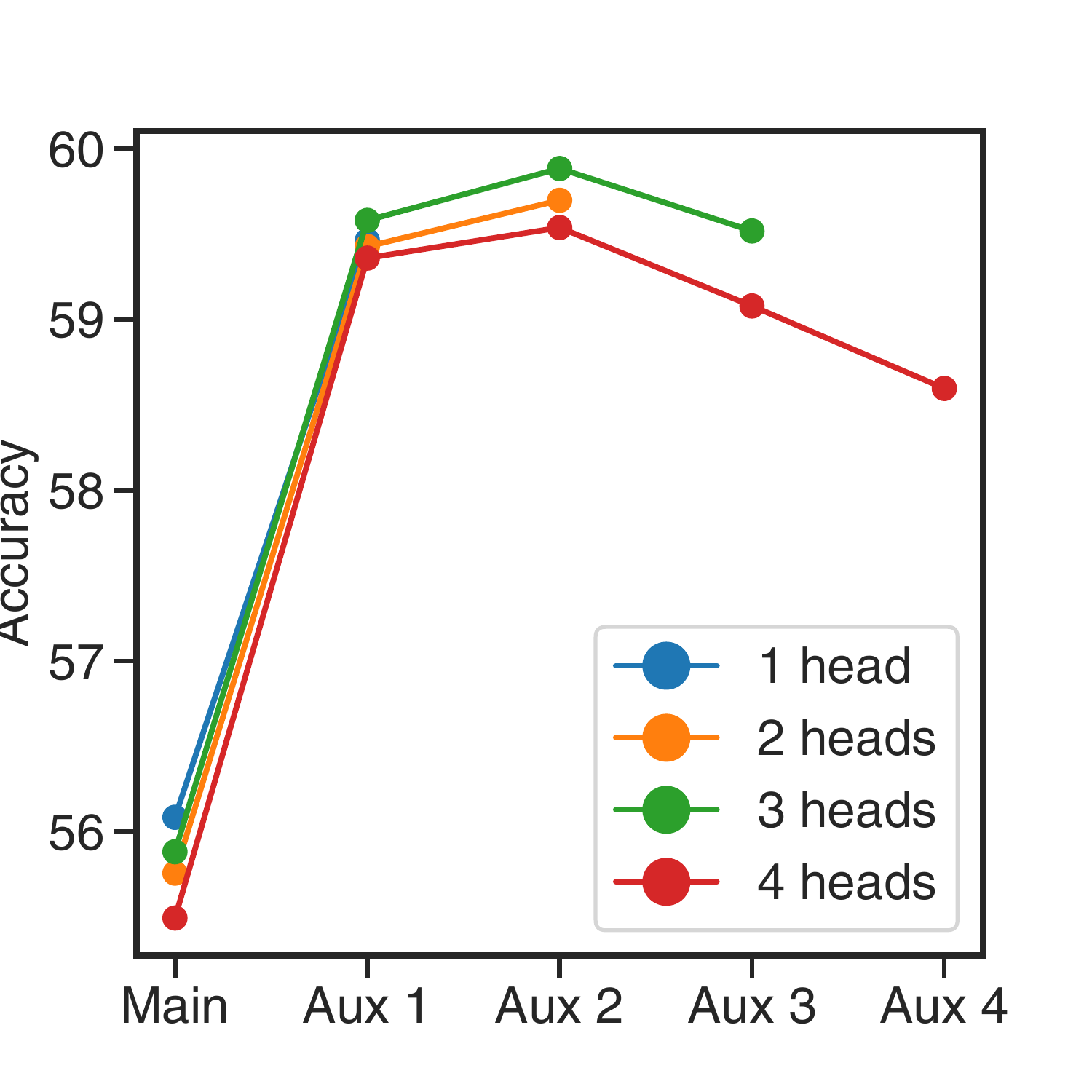}
            \caption{IID ($\skew=0$)}
            \end{subfigure}
            \begin{subfigure}[t]{0.23\textwidth}
            \includegraphics[width=0.97\textwidth]{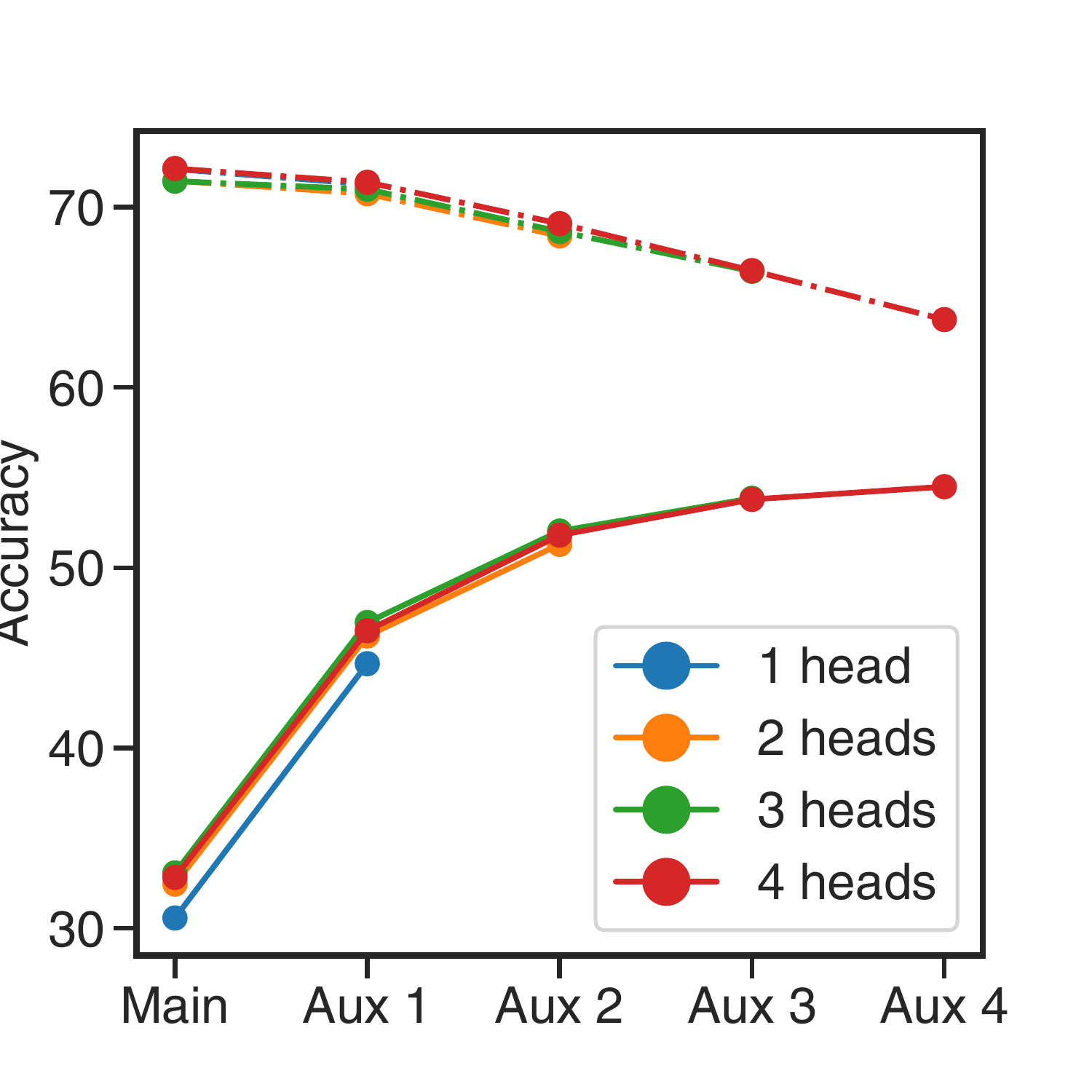}
            \caption{non-IID ($\skew=100$)}
            \end{subfigure}
            \caption{
                Private (dot-dashed) and shared (solid) dataset accuracies of main and auxiliary heads in ensembles trained with different numbers of auxiliary heads: $1$ aux head ({\color{NavyBlue} blue}), $2$ heads ({\color{Orange} orange}), $3$ heads ({\color{Green} green}) and $4$ heads ({\color{BrickRed} red}). For the IID case the private and shared performance match.
            }
            \label{fig:multiple-heads}
        \end{figure}
    
        \paragraph{Ignoring poor distillation targets.}
        In some cases, agents can be distilling knowledge about particular categories from agents that themselves do not possess accurate information.
        It is even possible that the agent's auxiliary head is already ``more knowledgeable'' about the class than the main head of another agent that it is trying to distill from.
        As a result, the performance of the auxiliary head may degrade.
        One approach that we study here is to skip distillation on a sample if the auxiliary head confidence is already higher than that of the head it is trying to distill from.
        In our experiments, we observed that that this simple idea had virtually no effect for $\skew=0$, but allowed us to improve the performance of the auxiliary head for heterogeneous data distributions with $\skew=100$.
        Specifically, for $8$ clients and $\skew=100$, this technique improved auxiliary head $\accsh$ from $44.7\%$ to $46.5\%$, while having virtually no effect on the private dataset accuracy $\accpriv$ of the main model head, which stayed at $72.2\%$.
        While effective for single auxiliary head, this technique did not improve  results in multiple auxiliary heads scenario (see Appendix~\ref{app:ablation}) that we will discuss next.
        
        \begin{table}
        	\small
            \begin{center} \begin{tabular}{cc|cc}
                $\skew=0$ & {\em Accuracy} & $\skew=100$ & {\em Accuracy} \\
                \hline \hline
                Separate & $46.3\%$ & Separate & $25.1\%$ \\
                \hline
                MHD (Ours) & $59.9\%$ & MHD (Ours) & $54.5\%$ \\
                MHD+ (Ours) & $68.6\%$ & MHD+ (Ours) & $63.4\%$ \\
                \hline
                FA, $u=200$ & $70.5\%$ & FA, $u=200$ & $68.0\%$ \\
                FA, $u=1000$ & $69.1\%$ & FA, $u=1000$ & $65.7\%$ \\
                \hline
                Supervised & $68.9\%$ & -- & -- \\
            \end{tabular} \end{center}
            \caption{
                \label{tab:results}
                Comparison of the shared accuracies $\accsh$ for our technique and two ``upper-bound'' baselines trained for $60{\rm k}$ steps on $90\%$ of \imagenet{}: (a) {\em supervised} and (b) trained with Federated Averaging ({\em FA}) performed every $u$ steps.
                {\em MHD+} experiments were conducted with $180{\rm k}$ steps and used the entire \imagenet{} as a public dataset (regime of plentiful public data).
                {\em Separate} corresponds to shared dataset performance for clients trained independently on their own private data.
                FA accuracy being higher than the supervised could be explained by a much larger number of samples being effectively processed during training ($\times 8$).
            }
        \end{table}    
        
        \paragraph{Multiple auxiliary heads.}
        Here we empirically study the multi-head approach inspired by self-distillation and described in detail in Section~\ref{sec:losses}.
        Guided by earlier results from Section~\ref{sec:exp-basic}, we choose $\regemb=1$ and $\regaux=3$.
        We then train an ensemble of $8$ models, each with $250$ primary labels and two choices of dataset skew: $\skew=0$ and $\skew=100$.
        For each choice of parameters, we independently trained models with $1$ to $4$ auxiliary heads and then measured the performance of the main and every auxiliary head on the client's private dataset and a shared test set with a uniform label distribution.
        The results of our experiments are presented in Figure~\ref{fig:multiple-heads} (see Appendix~\ref{app:ablation} for raw numbers).
        For a uniform data distribution, i.e., $\skew=0$, we see that distilling multiple auxiliary heads has a positive impact on all model heads for up to $3$ auxiliary heads, after which performance starts to degrade.
        Among the heads themselves, the peak performance is seen to be attained by the $2^{\rm nd}$ auxiliary head.
        However, we hypothesize that with the increase of the number of training steps, the final head will end up having the highest accuracy.
        
        In the case of a non-iid distribution with $\skew=100$, we observed that increasing the number of auxiliary heads has a very profound positive affect on the shared dataset performance $\accsh$ of the final auxiliary head.
        However, it is the main head that achieves the highest private dataset accuracy $\accpriv$.
        All consecutive auxiliary heads appear to loose their private dataset performance $\accpriv$ by specializing on capturing the overall data distribution.
    
        \paragraph{Dependence on the number of distillation targets $\numcheckpoints$.}
            We studied the effect of using multiple distillation targets $\numcheckpoints$ at every training step by considering a typical $8$-client setup with $\skew=100$, $4$ auxiliary heads, $\regemb=1$ and $\regaux=3$.
            While increasing $\numcheckpoints$ from $1$ to $3$ had virtually no effect on the main head private accuracy $\accpriv$, the shared dataset accuracy $\accsh$ for the last auxiliary head improved from $54.5\%$ to $56.1\%$ and then to $56.4\%$ as we increased $\numcheckpoints$ from $1$ to $3$.
            At $\numcheckpoints=4$, $\accsh$ appeared to saturate and fell to $56.2\%$ (within the statistical error of about $0.2\%$).
            Overall, earlier auxiliary heads appeared to be affected by $\numcheckpoints$ more strongly.
    
        \paragraph{Choice of the confidence measure.}
            The choice of the confidence $\confidence(\vec{h}(x))$ is central to the distillation technique.
            We compared our current choice based on selecting the most confident head, with a random selection of the distillation target.
            In our experiments with $8$ clients each with $250$ random primary labels, $\regemb=1$, $\regaux=3$, $\skew=0$ and $3$ auxiliary heads, we observed that randomizing confidence caused the main head $\accpriv$ degradation from $56\%$ to $55.2\%$ and the last auxiliary head $\accsh$ went down from $59.5\%$ to $58.4\%$.
            The degradation of model performance is more significant in the case of heterogeneous client data.
            In experiments with $\skew=100$ and $4$ auxiliary heads, we observed the main head $\accpriv$ degraded from $72.1\%$ to $71.3\%$ and the last auxiliary head $\accsh$ decreased from $54.5\%$ to $49\%$.
    
        \paragraph{Dependence on the technique efficiency on the public dataset size.}
        The efficiency of model distillation depends on the amount of data used for performing this distillation, in our case, on the size of the public dataset.
        In our experiments outlined in Appendix~\ref{app:public}, increasing the size of the public dataset while fixing the amount of private training data has a positive impact on the final model performance.
        
        In practice, since unlabeled data is more abundant, one can expect that the public dataset size will be comparable or even larger than the total amount of labeled data available to clients.
        Being constrained by the \imagenet{} size and attempting to keep the amount of private training data unaffected, we simulate the abundance of public data by reusing the entirety of the \imagenet{} dataset as an unlabeled public dataset.
        This, of course, is not realistic and somewhat biased given that we reuse the same samples as labeled and unlabeled, but it allows us to explore the limits of the distributed training efficiency with distillation.
    
        \begin{table}
        	\small
            \begin{center} \begin{tabular}{cccc}
                MHD Base & MHD & FedMD Base & FedMD \\
                \hline
                $60.6\%$ & $57.0\%$ / $0.6\%$ & $56.5\%$ & $50.2\%$ / $2.7\%$
            \end{tabular} \end{center}
            \caption{
                \label{tab:dist}
                Comparison of mean test accuracies (first number) and their deviations (second number after $/$) across $10$ clients for our method and FedMD as reported in Ref.~\cite{Li19}.
                Baselines (Base) are obtained by training clients with all available private data.
            }
        \end{table}
    
    \subsection{Baseline Comparisons}
    
        Before comparing our technique with a similar distillation-based method, we compared its performance with two strong ``upper-bound'' baselines (see Table~\ref{tab:results}): supervised training on all \imagenet{} and FedAvg algorithm implemented within our framework.
        A large performance gap between shared dataset accuracies obtained using our method and the strong baselines can be viewed as a price paid for learning via distillation in a decentralized multi-agent system.
        At the same time, we see that increasing the public dataset size and training for a longer period of time, allowing the information to propagate across all clients ({\em Our+} results), brings us close to the supervised model performance.
        Notice that like many other distillation-based techniques \cite{Li19,Zhang21}, our method reaches higher accuracy in the homogeneous data scenario.
        
        We compared our method with FedMD \cite{Li19} a similar, but {\em centralized} distillation-based methods.
        This comparison was carried out by replicating the dataset and $10$ model architectures from the publicly available implementation.
        The dataset is based on CIFAR-100 \cite{Krizhevsky2009} and makes use of $20$ coarse labels, while the public dataset is chosen to be CIFAR-10.
        Due to the differences in the training process, our baseline results with individual models trained on all private data pooled together was higher than that reported in \cite{Li19}.
        At the same time, we observed a much smaller gap in performance between this upper baseline and the results obtained using our method than the gap reported in \cite{Li19} (see Table~\ref{tab:dist}).
        Interestingly, we also observe a much smaller performance spread across all 10 models trained with our technique (deviation of $0.6\%$ compared to $2.7\%$ for FedMD).
        
    \subsection{Communication Topology Effects}
        \begin{figure}[b]
            \centering
            \includegraphics[width=0.45\textwidth]{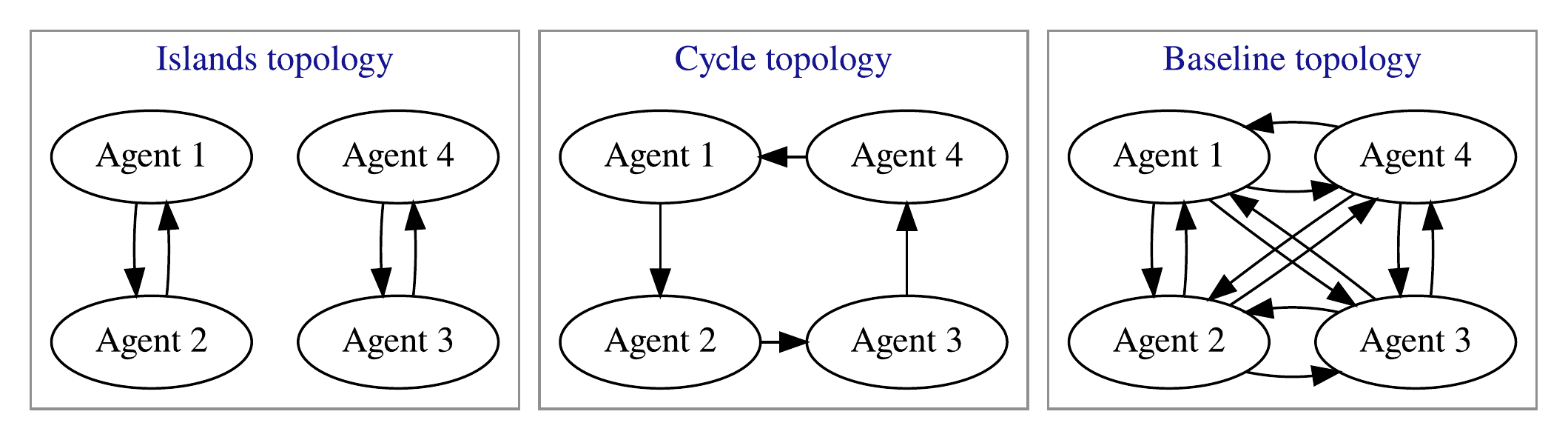}
            \caption{
                Topologies compared to validate transitive distillation.
            }
            \label{fig:topologies}
        \end{figure}
        \begin{figure}
            \centering
            \includegraphics[width=0.45\textwidth]{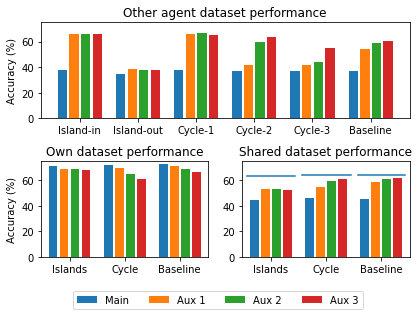}
            \caption{
                The performance by topology and distance between distillation teacher and student. On the shared dataset the blue horizontal lines indicate the upper bound per the embedding quality -- computed by fine tuning a head on the frozen model embeddings. Note that island embedding accuracy for ``Islands'' is still worse than ``Cycle''. Best viewed in color. 
            }
            \label{fig:topologies-perf}
        \end{figure}
        
        In order to explore how our approach might scale to larger systems in which pairwise connections between all agents are not feasible, we aim to evaluate how the communication topology affects performance. In particular we are interested in the question of whether ``transitive distillation'' is possible with our approach -- that is whether two agents that are not directly connected to one-another can still learn from each-other through an intermediary.
        
        To evaluate this and determine how auxiliary heads play a role in the performance we ran a training sweep with $4$ agents arranged in $3$ different topologies (Figure~\ref{fig:topologies}) with $3$ auxiliary heads each. In all cases we trained for $120{\rm k}$ steps, with $250$ primary labels per agent with $\skew=100$. We observe (Figure~\ref{fig:topologies-perf}) that performance on the shared dataset improves significantly between island and cycle topology, with the baseline performance matching closely the cycle performance. Without transitive distillation we would expect island and cycle performance to match closely so this provides strong evidence for transitive distillation. Also note that this behavior is only present on auxiliary heads and is more pronounced for later heads.
        
        We further analyze the performance of each agent on other agents' private data. Predictably we observe that island topologies perform well on in-island other agents, and poorly on agents from outside their island. Cycle topology agents perform best on their direct teacher ({\em Cycle-1}), but auxiliary heads 2 and 3 perform well on the ``1-hop'' transitive teacher ({\em Cycle-2}), and auxiliary head 3 has markedly improved performance on the ``2-hop'' transitive teacher ({\em Cycle-3}). We take this as strong evidence that auxiliary heads enable transitive distillation, and that additional heads make learning across additional degrees of separation more efficient.
    
    \subsection{Learning in Heterogeneous Systems}
        In Section~\ref{sec:exp-homo}, we conducted experiments with homogeneous ensembles of models.
        However, in many realistic scenarios of distributed deep learning, client devices may have different hardware-defined limitations and it may be desirable to train smaller models on some clients, while allowing other devices to utilize much larger networks.
        While model distillation allows one to achieve this, it is reasonable to ask why would this even be desirable?
        What do we expect to gain from having much larger models in the ensemble?
        Here we show two positive effects emerging from having larger models in an ensemble of smaller clients: (a) informally speaking, small models benefit from having stronger teachers and (b) large models can gain complex knowledge by distilling from smaller and simpler models.
        
        Our \imagenet{} experiments were conducted with $4$ clients each assigned $500$ primary labels with one client being a ResNet34 model and the remaining clients being ResNet18.
        Primary label assignment was random across clients and we trained the model for $240\textrm{k}$ steps.
        
        First, we observed that the presence of a larger model improved the accuracy of smaller clients suggesting that they benefited from seeing a stronger teacher holding some of the relevant data.
        Specifically, we observed that the presence of a ResNet34 model instead of ResNet18 in the ensemble led to an increase in the average shared accuracy $\accsh$ of ResNet18 models from $66.2\%$ to $66.7\%$.
        
        Secondly, if small models achieve high performance on their limited personalized domains, a large model distilling from such an ensemble can potentially learn a much more complex picture of the entire dataset than would otherwise be accessible to any individual small learner.
        This observation has already inspired centralized distillation-based methods like \cite{He20}.
        In our experiments, we witnessed this by observing that ResNet34 trained in conjunction with $3$ ResNet18 clients reached the shared accuracy $\accsh$ of $68.6\%$, which exceeds the $67.7\%$ accuracy of an ensemble of $4$ ResNet18 models trained with FedAvg or $66.0\%$ if trained with our approach (both with $200$ steps between updates).
        Notice that if the ResNet34 model is isolated from ResNet18 models, it only reaches $\accsh$ of $39.4\%$.

% =================================================================================
%   DISCUSSION AND CONCLUSIONS
% =================================================================================

\section{Discussion and Conclusions}

    In this paper, we proposed a novel distributed machine learning technique based on model distillation.
    The core idea of our approach lies in using a hierarchy of multiple auxiliary heads distilling knowledge from each other and across the ensemble.
    We show that this technique is much more effective than naive distillation and allows us to get close to the supervised accuracy on a large \imagenet{} dataset given a large public dataset and longer training time necessary for information to spread across the system.
    We also study two key capabilities of a distributed distillation-based learning technique.
    Specifically, we demonstrate that in systems where direct communication between the clients is limited, multiple auxiliary heads allow information exchange across clients that are not directly connected.
    We also demonstrate two positive effects of adding larger models into the system of small models: (a) small models benefit from seeing larger teachers and that (b) large models learning from a collection of small models can reach higher accuracies than those achievable with small models only.

{\small
\bibliographystyle{ieee_fullname}
\bibliography{paper}
}

\onecolumn
\newpage
\twocolumn
\appendix

\def\E{\mathbb{E}}
\def\R{\mathbb{R}}
\def\I{\mathbb{I}}
\def\H{\mathbb{H}}
\newcommand{\pd}[2]{\frac{\partial #1}{\partial #2}}
\newcommand{\mat}[1]{{\mathrm{#1}}}
\def\w{{\rm w}}

% =================================================================================
%   APPENDIX
% =================================================================================

\section{Analysis and Discussion of Our Method}
\label{app:theory}

    \subsection{Analysis of Multi-Headed Distillation}
    \label{app:self-dist-theory}
    
        As described in the main text, the multi-headed distillation involves simultaneous training of multiple model heads that communicate with each other.
        Rigorous theoretical analysis of this process in the most general case is very complicated.
        However, by making several assumptions, here we find an approximate solution for the weights of the heads of rank $k$ being given the weights of the heads of rank $k-1$.
        In our future work, we hope to study this model for different prediction aggregation techniques and compare conclusions obtained from this simple model with those obtained empirically in realistic systems.
        
        Let $X$ be the input space and $L=\R^d$ be the logit space, where $d$ is the number of classes.
        The logits $f(x)$ for a model $f:X\to L$ are then converted to label assignment probabilities $p(y|x)$ via softmax, i.e., $p(y|x) = {\rm softmax}(f(x))$.
        
        Consider a single client $h_i:X\to L$ distilling information from some model head $h:X\to L$.
        The corresponding distillation loss $\mc{L}[h_i;h]$ admits many possible choices, but we will assume that
        \begin{equation*}
            \mc{L} \equiv \E_{x\sim \mc{D}} \, D \left[ h_i(y|x;\psi_i) \, \middle\| \, p_{h}(y|x) \right]
        \end{equation*}
        with $\mc{D}$ being the shared (proxy) dataset and $D$ being some divergence (more general $f$-divergence or KL-divergence as some examples).
        The distillation is then carried out by performing optimization of $\mc{L}$, for example via gradient descent:
        \begin{equation*}
            \Delta \psi_i = - \gamma \pd{\mc{L}}{\psi_i}.
        \end{equation*}
        Notice that the components of $\psi_i$ corresponding to the model backbone may receive updates from multiple heads reusing the same model embedding.
        
        In our system, we assume that there is a set of heads $\{ h_i^{(1)}, h_i^{(2)}, \dots, h_i^{(n)} \}$ for each client $i$.
        For simplicity, let us first consider distillation procedure independent of prediction confidence.
        In this case, the loss for head $k$ of the client $i$ may look like:
        \begin{equation*}
            \mc{L}^{(k)}_i \equiv \sum_{j=1}^{N} \rho_{ij} \, \Gamma \left[ h^{(k)}_{i} \, \middle\| \, h_j^{(k-1)} \right],
        \end{equation*}
        where
        \begin{equation*}
            \Gamma \left[ h^{(k)}_{i} \, \middle\| \, h_j^{(k-1)} \right] \equiv \E_{x\sim \mc{D}} \, D \left[ p^{(k)}_i(y|x) \, \middle\| \, p_j^{(k-1)}(y|x) \right],
        \end{equation*}
        $p^{(k)}_i(y|x)$ is a shorter notation for $p_{h^{(k)}_i}(y|x)$ and $\rho_{ij}$ is some distribution defining the probability of picking a particular client for distillation.
        Again, here we assume that $\rho_{ij}$ does not depend on the sample confidence and is simply fixed.
        
        While we talked about $h^{(k)}$ distilling to $h^{(k-1)}$, we have not yet discussed the "main head" $h^{(1)}$.
        This head is normally trained locally on the client's private data.
        For simplicity, in the following we thus assume that its behavior is known, i.e., $h^{(1)}_i$ is a specified function of the training step.
        Furthermore, in the following, we start analyzing the problem by assuming that all $h_i^{(1)}$ already converged and are all generally different due to some differences in the client's private data.
        The behavior of all other heads is then defined by the losses outlined above.
        
        Let us first consider the simplest case of $\rho_{ij} = \delta_{ij}$.
        In other words, each head only distills from the same client's "prior" head.
        The choice of $h_i^{(n)} = \dots = h_i^{(1)}$ would obviously minimize all losses $\mc{L}_i^{(k)}$ since all corresponding $\Gamma[\cdot]$ values vanish.
        But as soon as we introduce a small correction $\rho_{ij} = \delta_{ij} + \nu_{ij}$ with $\sum_{j} \nu_{ij} = 0$, this trivial solution is no longer optimal.
        Instead, each client's head is now optimizing:
        \begin{equation*}
            \mc{L}^{(k)}_i = \Gamma \left[ h^{(k)}_{i} \, \middle\| \, h_{i}^{(k-1)} \right] + \sum_{j=1}^{N} \nu_{ij} \, \Gamma \left[ h^{(k)}_{i} \, \middle\| \, h_j^{(k-1)} \right].
        \end{equation*}
        
\begin{figure}[t]
            \centering
            \includegraphics[width=.24\textwidth]{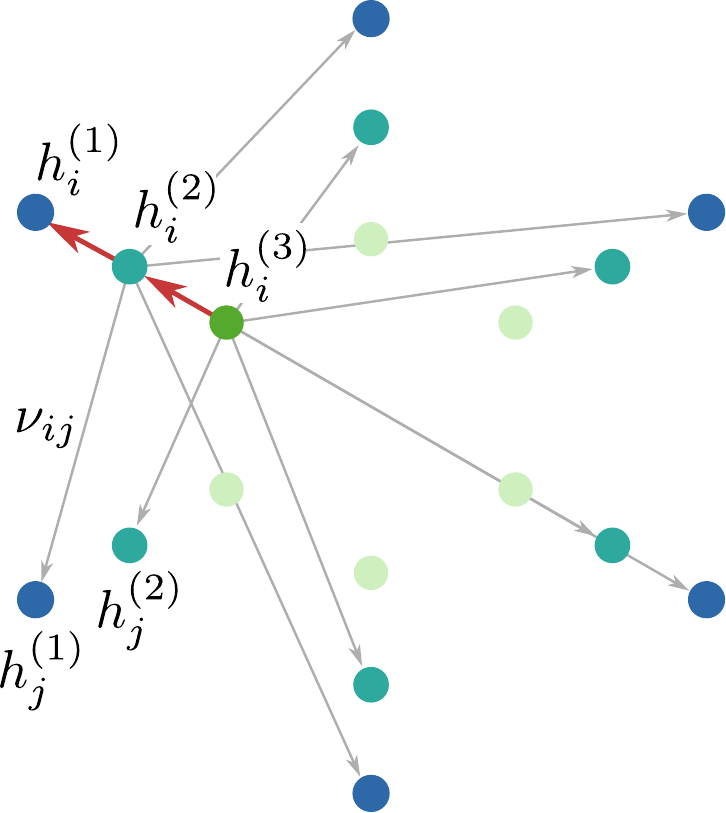}
            \caption{
                Illustration of multi-head distillation as discussed in Appendix~\ref{app:self-dist-theory}.
                Large red arrow shows strong distillation of $h_i^{(k)}$ to $h_i^{(k-1)}$ and smaller gray arrows indicate attraction towards $h_j^{(k-1)}$ with effective ``strength'' $\nu_{ij}$.
            }
            \label{fig:h-propagation}
        \end{figure}

        Notice that if $\Gamma$ was a metric in the $h$ space, we could interpret this optimization objective geometrically as a minimization of the head's distance to its lower-order state (towards $h_i^{(k-1)}$) coupled with a weak ($\sim\nu$) attraction towards a number of other heads ($h_j^{(k-1)}$).
        See Figure~\ref{fig:h-propagation} for illustration.
        
        Here we have to make yet another simplifying assumption and consider a prescribed model backbone (and corresponding embedding) that we are not optimizing or updating with backpropagated gradients.
        Doing so, we disentangle individual heads and can treat their optimization as independent tasks.
        For sufficiently small $\nu_{ij}$ it will hold that $p_i^{(k)} = p_i^{(k-1)} + O(\nu)$ and we can therefore write:
        \begin{multline*}
            \mc{L}^{(k)}_i = \E_{x\sim \mc{D}} \Biggl\{ D \left[ \left. p_{h^{(k-1)}_{i} + \kappa^{(k)}_i} \, \right\| \, p_{h_{i}^{(k-1)}} \right] + \\ +
            \sum_{j=1}^{N} \nu_{ij}  D \left[ \left. p_{h^{(k-1)}_{i} + \kappa^{(k)}_i} \, \right\| \, p_{h_j^{(k-1)}} \right] \Biggr\},
        \end{multline*}
        where $h_i^{(k)} \equiv h_{i}^{(k-1)} + \kappa^{(k)}_i$ and $\kappa^{(k)}_i \sim O(\nu)$.
        Introducing $\delta_i^{(k)} \equiv p_{i}^{(k)} - p_{i}^{(k-1)}$, we obtain:
        \begin{multline*}
            \mc{L}^{(k)}_i =
            \E_{x\sim \mc{D}} \Biggl\{ D \left[ \left. p^{(k-1)}_{i} + \delta_i^{(k)} \, \right\| \, p_{i}^{(k-1)} \right] + \\ +
            \sum_j \nu_{ij} \, D \left[ \left. p^{(k-1)}_{i} + \delta_i^{(k)} \, \right\| \, p_{j}^{(k-1)} \right]
            \Biggr\}.
        \end{multline*}
        Noticing that the first term needs to be decomposed near the global minimum and the second term permits linear expansion, we obtain:
        \begin{multline*}
            \mc{L}^{(k)}_i \approx 
            \E_{x\sim \mc{D}} \Biggl\{ D'' \left[ \left. p^{(k-1)}_{i} \, \right\| \, p_{i}^{(k-1)} \right] \frac{\delta_i^{(k)} \delta_i^{(k)}}{2} + \\ +
            \sum_j \nu_{ij} \, D' \left[ \left. p^{(k-1)}_{i} \, \right\| \, p_{j}^{(k-1)} \right] \delta_i^{(k)}
            \Biggr\},
        \end{multline*}
        where $D''$ and $D'$ are the derivatives of $D$ with respect to the first argument.
        Recalling that $\delta_i^{(k)} \in \R^d$ we can rewrite the loss function as:
        \begin{equation*}
            \mc{L}^{(k)}_i \approx 
            \E_{x\sim \mc{D}} \left[ \delta^\top\mat{A} \delta + 
            b^\top \delta
            \right],
        \end{equation*}
        where $\delta \equiv \delta_i^{(k)}$ for brevity,
        \begin{equation*}
            \mat{A} \equiv D'' \left[ \left. p^{(k-1)}_{i} \, \right\| \, p_{i}^{(k-1)} \right] / 2
        \end{equation*}
        is effectively a matrix and
        \begin{equation*}
            b \equiv \sum_j \nu_{ij} \, D' \left[ \left. p^{(k-1)}_{i} \, \right\| \, p_{j}^{(k-1)} \right] \in \R^d
        \end{equation*}
        can be thought of as a column vector.
        
        At this point we can connect the probability distribution perturbation $\delta$ to the logit perturbation $\kappa \equiv \kappa^{(k)}_i$ using the fact that $p_m \equiv e^{h_m}/Z$, where $Z \equiv \sum_k e^{h_k}$ (we omit this simple calculation here):
        \begin{multline*}
            p_i^{(k)} = p_{h_i^{(k-1)} + \kappa_i^{(k)}} = p_i^{(k-1)} + \delta = \\ = p_i^{(k-1)} + \kappa*p_i^{(k-1)} - (\kappa \cdot p_i^{(k-1)}) p_i^{(k-1)},
        \end{multline*}
        where $\vec{a}*\vec{b}$ is an element-wise product of two vectors and therefore:
        \begin{equation}
            \label{eq:c-def}
            \delta = \kappa*p_i^{(k-1)} - (\kappa \cdot p_i^{(k-1)}) p_i^{(k-1)} \equiv \mat{C} \kappa,
        \end{equation}
        where $\mat{C}$ is a matrix constructed from the components of $p_i^{(k-1)}(x) \in \R^d$.
        Notice that $\sum_m \delta_m = 0$, which agrees with $\delta$ being the perturbation of the normalized probability distribution.
        
        Finally, remember that $\kappa$ itself is a perturbation of model logits.
        Given the sample embedding $\xi_i(x) \in \R^t$, the sample logits are constructed as $\mat{W}_i \xi_i(x)$ with $\mat{W}_i$ being a $d\times t$ matrix.
        The perturbation $\kappa$ transforming $\mat{W}_i^{(k-1)}\xi_i$ into $\mat{W}_i^{(k)}\xi_i$ can thus be characterized by the logit weight perturbation $\mu \equiv \mu_i^{(k)} := \mat{W}_i^{(k)} - \mat{W}_i^{(k-1)}$ and we get $\kappa = \mu \xi(x)$.
        Combining everything together, we see that the loss function transforms to:
        \begin{equation}
            \label{eq:fin-loss}
            \mc{L}^{(k)}_i \approx 
            \E_{x\sim \mc{D}} \left[ \xi(x)^\top \mu^\top \mat{C}^\top \mat{A} \mat{C} \mu \xi(x) + 
            b^\top \mat{C} \mu \xi(x)
            \right],
        \end{equation}
        where $\mat{A}$, $\mat{C}$ and $b\sim \nu$ all depend on the sample $x$ via $p_i^{(k-1)}(x)$ and $\xi$ is a function of $x$, while $\mu$ is effectively an unknown sample-independent matrix that we need to tune with the goal of minimizing $\mc{L}^{(k)}_i$.
        The optimum can be identified by taking a derivative with respect to $\mu_{\alpha\beta}$ and setting it to 0:
        \begin{equation*}
            \E_{x\sim \mc{D}} \left[ 2 (\xi(x)^\top \mu^\top \mat{C}^\top \mat{A} \mat{C})_{\alpha} \xi_\beta(x) + 
            (b^\top \mat{C})_{\alpha} \xi_\beta(x)
            \right] = 0.
        \end{equation*}
        This is a linear equation on $\mu \sim \nu$ that can be solved in a closed form to give us a logit weight perturbation $\mu$ as a complex nonlinear function of $\nu_{ij}$ and $\{ p_\ell^{(k-1)} \}$.
        
        Note that since $\mu$ is only a small perturbation, we can introduce $\mat{W}_i^{(k)}$ as a function of a {\em continuous} parameter $k$ and approximate $d\mat{W}_i^{(k)} / dk$ with a finite difference $\mat{W}_i^{(k)} - \mat{W}_i^{(k-1)} = \mu$ leaving us with a differential equation (the approximation is valid in the first order in $\nu$):
        \begin{equation*}
            \frac{d\mat{W}_i(k)}{dk} = G\left[ \nu, \{\mat{W}_\ell(k)\} \right]
        \end{equation*}
        with $G$ being a linear function with respect to $\nu$, but very complex nonlinear function with respect to $\{ \mat{W}_\ell \}$.
        If $\nu_{ij}$ is localized around $i=j$ (which would be the case for communication patterns with partial connectivity, like in the case of long chains), this differential equation resembles a complex nonlinear diffusion equation defining the spread of information across the clients as we look at deeper and deeper heads (with the head rank $k$ essentially playing the role of time).
        
        It is also worth noting here that if $\nu$ was not fixed, but was itself a function of model confidence (while still remaining small), our conclusions would not change except that $\nu$ itself would now itself be a complex nonlinear function of $\{\mat{W}_\ell(k)\}$ and $x$.
        In our future work, we hope to study the effect that this confidence-dependent aggregation has on head dynamics and the final stationary state.
        
        Finally, let us look at the stationary state of system dynamics.
        Equation \eqref{eq:fin-loss} suggests that $\mu=0$ is a local optimum when $b^\top \mat{C} = 0$, or
        \begin{gather*}
            \sum_{i,j} \nu_{ij} D'\left[ p_i \| p_j \right] \mat{C}_{ik} = 0,
        \end{gather*}
        or after noticing that $D'[ p_i \| p_j ] = p_j / p_i$ and recalling that $\mat{C}$ is defined by Eq.~\eqref{eq:c-def} we obtain for every $k$:
        \begin{gather}
            \label{eq:stat}
            \E_{x\sim \mc{D}} \left[ \sum_{i,j} \nu_{ij} p_j \left( \delta_{ik} - p_k \right) \right] = 0.
        \end{gather}
        Since $\sum_{j} \nu_{ij} = 0$, the trivial solution of this system of equations is the case of identical models, i.e., $p_{1}=\dots=p_{n}$, but since generally the models might have different embeddings and cannot be made identical, the solution of Eq.~\eqref{eq:stat} restricts the system stationary state.
    
    \subsection{Value of $p(y|x)$ as Classifier Confidence}
    \label{sec:know-agg}
    
        In our model distillation approach, we need to combine predictions of multiple different model heads.
        If all predictions $p_k(y|x)$ (by heads $\{h_k\}$) come with reliable error estimates, this information can be taken into account.
        For example, if we know that for the true distribution $p(y|x)$ and every prediction $p_k(y|x)$ it holds that $D[p_k(y|x) \| p(y|x)] \le e_k(x)$ with $D$ being some divergence, the true $p(y|x)$ belongs to the intersection of ``balls''\footnote{note that $D$ is not generally a metric} $\mc{B}_k \equiv \{ p' | D[p' \| p] \le e_k\}$.
        We can then choose any point in this set and compute a prediction error as a maximum distance from a chosen distribution to any point in the intersection.
        Unfortunately, however such reliable measures of classifier error are not generally available and even approximating them can be quite difficult.
        
        Instead we choose a very simple approach based on estimating classifier confidence and picking the most confident model, effectively ignoring other predictions.
        The confidence measure itself is chosen as a value of the largest component of the classifier prediction $o(x) \equiv \mathrm{softmax}(f(x;\theta))$ with $f(x;\theta)=\mat{W}\xi(x;\theta)$ and $\xi$ being the embedding vector.
        
        \paragraph{Why choose $\max_k o_k$.}
        This value has a number of trivial properties that can actually make it a useful measure of classifier uncertainty.
        First is that after seeing a supervised training sample $(x,y)$, the value of $o_y(x)$ is increased.
        Second is that if the class prototypes in the embedding space are nearly orthogonal for different classes, then updates for samples of different classes would not ``interfere'' with each other and high-confidence predictions would not generally be disrupted by observing unrelated samples with different labels.
        For a simple logits layer $\mat{W}\xi(x)$ trained with cross-entropy loss, both of these properties trivially follow from the following expression for $\Delta o_k(x')$ after training on a sample $(x,y)$:
        \begin{multline*}
            \Delta o_k(x') =
            \lambda o_k(x') [\xi(x) \cdot \xi(x')] \times \\ \times \sum_{i} \left( \delta_{k,i} - o_i(x') \right) \left( \delta_{y,i} - o_i(x) \right).
        \end{multline*}
        
        \paragraph{Drawbacks.}
        But while $\max_k o_k(x)$ has these useful properties, it is not guaranteed to be a reliable measure of classifier confidence for out-of-distribution samples and the training objective never explicitly optimizes for that\footnote{Contrast this to the energy-based models, for example, where the energy update and the MCMC sampling are explicitly contributing to the model ``awareness'' of what in-distribution samples are and are not.}.
        A density model $\rho(x)$ would allow detecting such out-of-distribution samples, but could also reveal information about the client samples in their private dataset.
        Combining classification models with locally-trained density models, or adopting other existing similar techniques could be a logical extension of our present work.
    
    \subsection{Distillation as Revelation of Some Information about Model Weights}
    
        The canonical version of FedAvg combines the knowledge of individual clients by periodically aggregating their weight snapshots.
        Distillation-based techniques are instead based on communicating model predictions on datasets accessible to all participants.
        While these two approaches appear to be different, communication in distillation-based methods can of course also be viewed as a way of revealing incomplete information about model weights.
        
        The amount of revealed information can be defined as follows.
        Assuming the knowledge of the prior $p(\theta)$ on the model weights and model predictions $(y_1,\dots,y_n)$ on a public dataset $\mc{D}_*=(x_1,\dots,x_n)$, one can compare the difference of entropies for the original $p(\theta)$ and $p(\theta|y_1,\dots,y_n)$ with
        \begin{gather*}
            p(\theta|y_1,\dots,y_n) = \frac{ p(y_1,\dots,y_n|\theta) p(\theta) } { \int d\theta \, p(y_1,\dots,y_n|\theta) p(\theta) }.
        \end{gather*}
        While generally intractable, it might be possible to obtain the lower bound on the amount of the revealed information by training a model that predicts the weights $\theta$ from $(y_1,\dots,y_n)$.
    
        Deeper understanding of this question can have an impact on the optimal choice of the public dataset $\mc{D}_*$ that would allow us to retrieve the knowledge of interest from a trained model using only a small number of samples.
        Ongoing research on dataset distillation \cite{Cazenavette22,Nguyen21,Wang18} is very closely related to this question.
    
\section{Additional Experiments and Experimental Data}
\label{app:ablation}
    
    \subsection{Effect of Distilling to Self and Same-Level Heads}
    
\begin{figure}[t]
        \centering
        \includegraphics[width=0.38\textwidth]{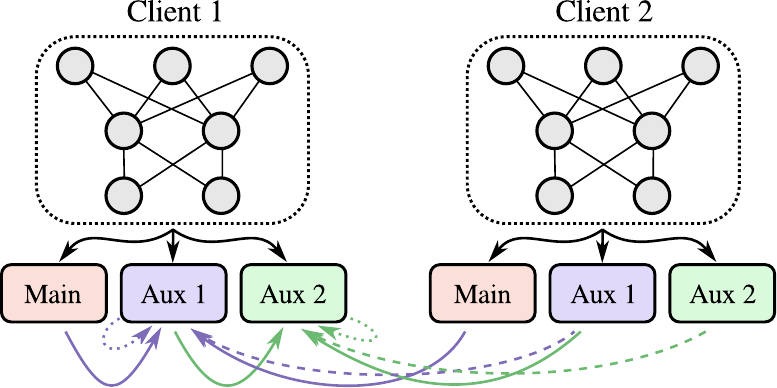}
        \caption{
            A pattern used for distilling multiple auxiliary heads with two additional types of distillation targets: (a) distilling to heads of the same ``rank'' (dashed), (b) distilling to ``self'' (dotted).
            Here distilling to the same ``rank'' means, for example, that {\em Aux1} head is distilled to the most confident of {\em Main} heads, or {\em Aux1} heads of adjacent clients.
            Distilling to ``self'' means that the samples on which the distilled head is already most confident will effectively be ignored.
        }
        \label{fig:aux-ext}
    \end{figure}

        In Section~\ref{sec:improving} we reported that including a head into the list of its own distillation targets (``self'') improved the model accuracy, but the gain was still smaller than that of a model with multiple auxiliary heads.
        Here we explore what happens if we use the head as its own potential distillation target, while also using a number of auxiliary heads.
        Furthermore, what if we modify our method to include distillations to other heads of the same rank (see Figure~\ref{fig:aux-ext})?
        
        We conducted a set of experiments with a heterogeneous dataset with $\skew=100$, $\regemb=1$, $\regaux=3$, four auxiliary heads and $250$ randomized labels per each of $8$ clients.
        The results of experiments using different combinations of distillation targets and both $\Delta=1$ and $\Delta=2$ (choosing two other clients at a time as potential distillation targets) are presented in Table~\ref{tab:head-vars}.
        We observed that using same-level heads and ``self'' targets {\em separately} provides noticeable benefit only for earlier heads.
        But when used together, these two techniques result in $\sim 1\%$ accuracy improvement and this improvement is realized for the $2^{\rm nd}$ auxiliary head.
        Also, not unexpectedly, using two clients to distill to ($\Delta=2$) instead of one, leads to a noticeable $1.5\%$ accuracy improvement.
        Combined together, all these techniques, in conjunction with using the entire \imagenet{} as the public dataset improve the accuracy to $59.4\%$ if trained for $60{\rm k}$ steps, or $65.7\%$ if trained for $180{\rm k}$ steps.
    
        \begin{table*}
            \centering
            \begin{tabular}{l|c|cccc}
                Experiment & $\accpriv^{\rm Main}$ & $\accsh^{\rm Aux1}$ & $\accsh^{\rm Aux2}$ & $\accsh^{\rm Aux3}$ & $\accsh^{\rm Aux4}$ \\
                \hline
                Base & $70.9\%$ & $46.7\%$ & $51.8\%$ & $53.9\%$ & ${\bf 54.6\%}$ \\
                $\Delta=2$ & $71.1\%$ & $50.9\%$ & $55.1\%$ & ${\bf 56.1\%}$ & ${\bf 56.0\%}$ \\
                SL & $70.8\%$ & $48.6\%$ & $53.6\%$ & ${\bf 54.7\%}$ & ${\bf 54.7\%}$ \\
                SF & $71.3\%$ & $48.1\%$ & $53.4\%$ & ${\bf 54.9\%}$ & $\bf{54.8\%}$ \\
                SL+SF & $70.3\%$ & $53.0\%$ & ${\bf 55.5\%}$ & $53.9\%$ & $52.4\%$ \\
                All & $70.8\%$ & $53.5\%$ & ${\bf 55.8\%}$ & $54.5\%$ & $52.9\%$ \\
                All+ & $72.7\%$ & $56.5\%$ & ${\bf 59.4\%}$ & $57.9\%$ & $56.1\%$ \\
                All+, $180{\rm k}$ steps & $76.2\%$ & $62.3\%$ & ${\bf 65.7\%}$ & $65.0\%$ & $64.0\%$ \\
            \end{tabular}
            \caption{
                Experimental results exploring the usage of different distillation heads trained for $60{\rm k}$ steps.
                Here ``{\em Base}'' is the original experiment with $\Delta=1$ and conventional heads as described in Sec.~\ref{sec:improving}; ``{\em SL}'' adds same-level heads to distillation targets; ``{\em SF}'' adds the distilled head (``self'') as a potential target; ``{\em All}'' combines same-level and ``self'' heads and $\Delta=2$ (each step distilling to two other clients), ``{\em All+}'' is the same as {\em All}, but also uses the entire \imagenet{} as the public dataset.
            }
            \label{tab:head-vars}
        \end{table*}
    
    \subsection{Dependence on the Public Dataset Size}
    \label{app:public}
        In a separate set of experiments, we trained $8$ clients with 4 auxiliary heads, $\skew=100$, $\regemb=1$, $\regaux=3$ and $250$ randomly assigned ``private'' labels and ``private'' samples drawn from $70\%$ of the \imagenet{} training data.
        The remaining $30\%$ of \imagenet{} samples were fully or partly used as a public dataset, i.e., $\fraction{\rm pub} \le 30\%$.
        As one would expect, increasing the size of the ``public'' dataset while fixing the amount of ``private'' training data has a positive impact on the final model performance (see Table~\ref{tab:fraction}).
    
        \begin{table}
        	\small
            \begin{center} \begin{tabular}{l|cccc}
                {\em Public DS fraction} & $10\%$ & $20\%$ & $30\%$ & {\em All} \\
                \hline
                {\em main head $\accpriv$} & $70.1\%$ & $71.1\%$ & $70.9\%$ & $71.9\%$ \\
                {\em last aux head $\accsh$} & $52.4\%$ & $53.9\%$ & $54.1\%$ & $55.3\%$ 
            \end{tabular} \end{center}
            \caption{
                \label{tab:fraction}
                The dependence of the main head ``private'' accuracy $\accpriv$ and the ``shared'' accuracy of the $4^{\rm th}$ auxiliary head on the size of the public dataset (fraction of \imagenet{} training set).
        		Experiments were conducted for a system of $8$ clients with 4 auxiliary heads, $\skew=100$, $\regemb=1$, $\regaux=3$ and $250$ randomly assigned ``private'' labels.
        		Private training samples were drawn from $70\%$ of the \imagenet{} training set in all experiments.
        		``{\em All}'' column shows the accuracy attained by using the entire \imagenet{} training set as a public dataset (while still using only $70\%$ of it as private data).
            }
        \end{table}
    
\section{Additional Tables and Figures}
    
    \begin{figure}[t]
        \centering
        \includegraphics[width=0.45\textwidth]{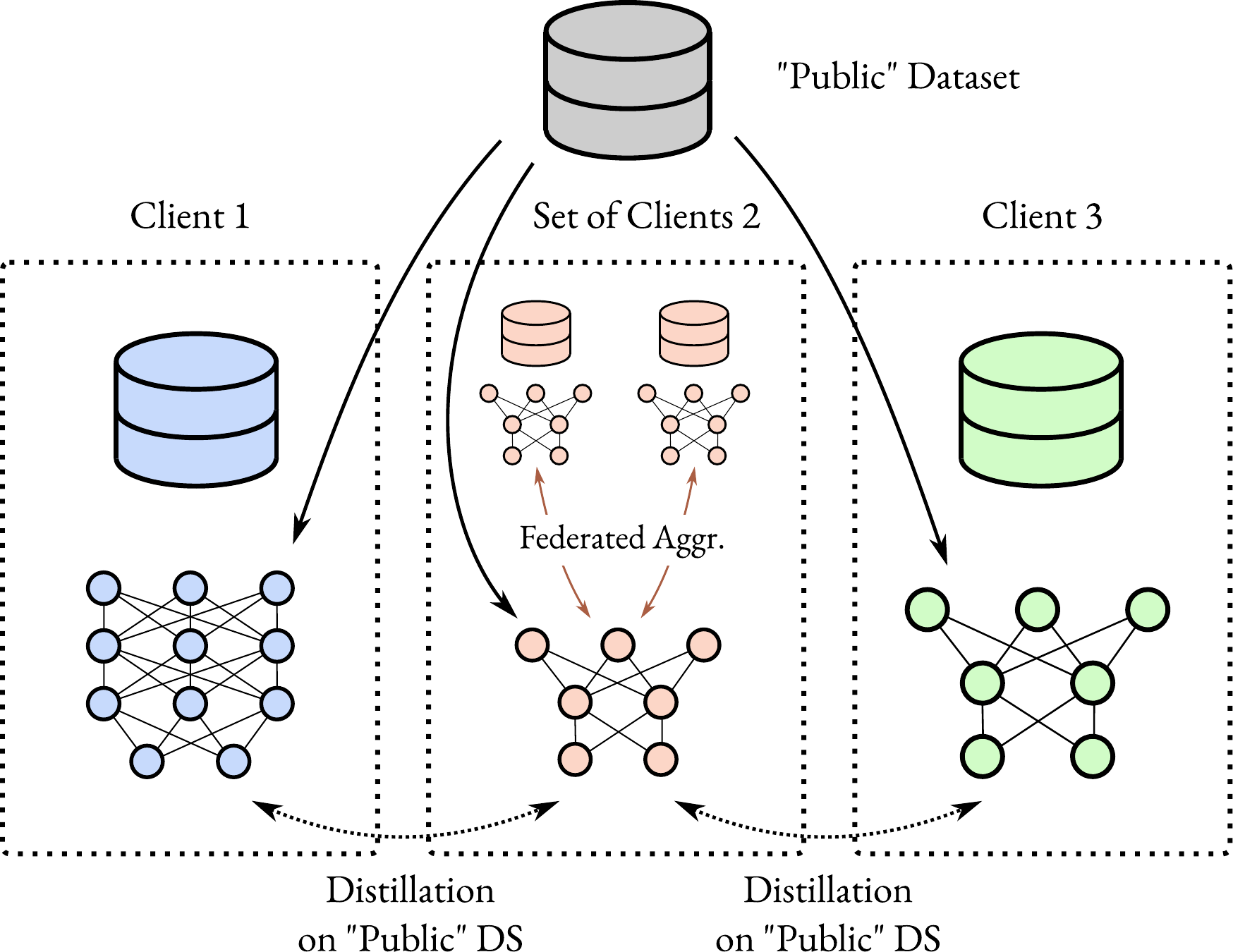}
        \caption{
            Conceptual diagram of a distillation in a distributed system.
            Clients use a ``public'' dataset to distill knowledge from other clients, each having their primary private dataset.
            Individual clients may have different architectures and different objective functions.
            Furthermore, some of the ``clients'' may themselves be collections of models trained using federated learning.
        }
        \label{fig:main_updated}
    \end{figure}

    \subsection{Raw Experimental Data}
    
        Tables~\ref{tab:skew_0} and \ref{tab:skew_100} contain raw values used for producing Figure~\ref{fig:regularization-effect}, while Tables~\ref{tab:heads_skew_0} and \ref{tab:heads_skew_100} complement Figure~\ref{fig:multiple-heads}.
        
        \begin{table}
            \centering
            \begin{tabular}{ll|cccc}
                $\regemb$ & $\regaux$ & $\accpriv^{\rm (m)}$ & $\accsh^{\rm (m)}$ & $\accpriv^{\rm (aux)}$ & $\accsh^{\rm (aux)}$ \\
                \hline
                0.0 & 0.0 & $46.3\%$ & $46.3\%$ & $0.1\%$ & $0.1\%$ \\
                & 1.0 & $52.2\%$ & $52.0\%$ & $56.0\%$ & $56.0\%$ \\
                & 3.0 & $54.1\%$ & $53.5\%$ & $57.3\%$ & $57.0\%$ \\
                & 10.0 & $54.3\%$ & $54.1\%$ & $57.1\%$ & $57.1\%$ \\
                \hline
                1.0 & 0.0 & $48.5\%$ & $48.5\%$ & $0.1\%$ & $0.1\%$ \\
                & 1.0 & $53.6\%$ & $53.5\%$ & $57.3\%$ & $57.2\%$ \\
                & 3.0 & $55.4\%$ & $55.2\%$ & $58.6\%$ & $58.5\%$ \\
                & 10.0 & $54.1\%$ & $53.6\%$ & $56.9\%$ & $56.3\%$ \\
                \hline
                3.0 & 0.0 & $48.3\%$ & $48.0\%$ & $0.1\%$ & $0.1\%$ \\
                & 1.0 & $54.3\%$ & $54.0\%$ & $58.2\%$ & $57.6\%$ \\
                & 3.0 & $55.7\%$ & $55.5\%$ & $59.3\%$ & $58.8\%$ \\
                & 10.0 & $53.3\%$ & $53.4\%$ & $56.5\%$ & $56.3\%$
            \end{tabular}
            \caption{
                Results for $8$-client experiments with $250$ random classes per client, $\skew=0$ and a varying values of $\regemb$ and $\regaux$.
            }
            \label{tab:skew_0}
        \end{table}
        
        \begin{table}
            \centering
            \begin{tabular}{ll|cccc}
                $\regemb$ & $\regaux$ & $\accpriv^{\rm (m)}$ & $\accsh^{\rm (m)}$ & $\accpriv^{\rm (aux)}$ & $\accsh^{\rm (aux)}$ \\
                \hline
                0.0 & 0.0 & $68.0\%$ & $25.2\%$ & $0.1\%$ & $0.1\%$ \\
                & 1.0 & $70.6\%$ & $29.1\%$ & $70.5\%$ & $42.0\%$ \\
                & 3.0 & $70.9\%$ & $30.0\%$ & $70.1\%$ & $43.3\%$ \\
                & 10.0 & $68.0\%$ & $25.3\%$ & $66.0\%$ & $39.7\%$ \\
                \hline
                1.0 & 0.0 & $69.0\%$ & $26.0\%$ & $0.1\%$ & $0.1\%$ \\
                & 1.0 & $71.8\%$ & $29.8\%$ & $71.5\%$ & $43.0\%$ \\
                & 3.0 & $72.0\%$ & $29.9\%$ & $71.0\%$ & $44.1\%$ \\
                & 10.0 & $66.8\%$ & $23.0\%$ & $65.1\%$ & $37.5\%$ \\
                \hline
                3.0 & 0.0 & $65.2\%$ & $24.9\%$ & $0.1\%$ & $0.1\%$ \\
                & 1.0 & $71.7\%$ & $29.7\%$ & $72.1\%$ & $39.1\%$ \\
                & 3.0 & $71.8\%$ & $29.7\%$ & $71.9\%$ & $40.8\%$ \\
                & 10.0 & $65.4\%$ & $23.1\%$ & $63.4\%$ & $36.4\%$
            \end{tabular}
            \caption{
                Results for $8$-client experiments with $250$ random classes per client, $\skew=100$ and a varying values of $\regemb$ and $\regaux$.
            }
            \label{tab:skew_100}
        \end{table}
        
        \begin{table}
            \centering
            \begin{tabular}{c|cccc}
                Heads & $1$ & $2$ & $3$ & $4$ \\
                \hline
                $\accpriv^{\rm (m)}$ & $56.2\%$ & $56.1\%$ & $55.8\%$ & $55.9\%$ \\
                $\accsh^{\rm (m)}$ & $56.1\%$ & $55.8\%$ & $55.8\%$ & $55.5\%$ \\
                $\accpriv^{\rm (1)}$ & $59.6\%$ & $59.6\%$ & $59.4\%$ & $59.4\%$ \\
                $\accsh^{\rm (1)}$ & $59.4\%$ & $59.5\%$ & $59.6\%$ & $59.4\%$ \\
                $\accpriv^{\rm (2)}$ &  & $60.0\%$ & $59.7\%$ & $59.7\%$ \\
                $\accsh^{\rm (2)}$ &  & $59.7\%$ & $59.9\%$ & $59.5\%$ \\
                $\accpriv^{\rm (3)}$ &  &  & $59.5\%$ & $59.1\%$ \\
                $\accsh^{\rm (3)}$ &  &  & $59.5\%$ & $59.1\%$ \\
                $\accpriv^{\rm (4)}$ &  &  &  & $58.7\%$ \\
                $\accsh^{\rm (4)}$ &  &  &  & $58.6\%$ \\
            \end{tabular}
            \caption{
                Results for $8$-client experiments with $250$ random classes per client, $\skew=0$, $\regemb=1$, $\regaux=3$ and a varying number of auxiliary heads (separate columns).
            }
            \label{tab:heads_skew_0}
        \end{table}
        
        \begin{table}
            \centering
            \begin{tabular}{c|cccc}
                Heads & $1$ & $2$ & $3$ & $4$ \\
                \hline
                $\accpriv^{\rm (m)}$ & $72.5\%$ & $71.6\%$ & $71.1\%$ & $72.5\%$ \\
                $\accsh^{\rm (m)}$ & $30.5\%$ & $32.5\%$ & $33.1\%$ & $32.7\%$ \\
                $\accpriv^{\rm (1)}$ & $71.4\%$ & $70.6\%$ & $70.7\%$ & $71.4\%$ \\
                $\accsh^{\rm (1)}$ & $44.7\%$ & $46.6\%$ & $46.9\%$ & $46.4\%$ \\
                $\accpriv^{\rm (2)}$ &  & $68.5\%$ & $68.1\%$ & $68.7\%$ \\
                $\accsh^{\rm (2)}$ &  & $51.6\%$ & $52.0\%$ & $51.6\%$ \\
                $\accpriv^{\rm (3)}$ &  &  & $66.1\%$ & $66.1\%$ \\
                $\accsh^{\rm (3)}$ &  &  & $53.8\%$ & $53.6\%$ \\
                $\accpriv^{\rm (4)}$ &  &  &  & $63.4\%$ \\
                $\accsh^{\rm (4)}$ &  &  &  & $54.5\%$
            \end{tabular}
            \caption{
                Results for $8$-client experiments with $250$ random classes per client, $\skew=100$, $\regemb=1$, $\regaux=3$ and a varying number of auxiliary heads (separate columns).
            }
            \label{tab:heads_skew_100}
        \end{table}

\end{document}